\documentclass[acmtog,screen]{acmart}

\acmSubmissionID{0473}

\acmJournal{TOG}
\acmYear{2022}\acmVolume{41}\acmNumber{4}\acmArticle{166}\acmMonth{7} \acmDOI{10.1145/3528223.3530130}

\ccsdesc[500]{Computing methodologies~Rendering}
    
\keywords{Relighting, NeRF, Eye Modeling, Regazing, Neural Rendering, Novel View Synthesis, Pose Optimization, Model Fitting, Volumetric Rendering, Differentiable Rendering, Refraction, HDR rendering, specularity synthesis, geometry deformation modeling}

\setcopyright{rightsretained}

\usepackage[utf8]{inputenc}
\usepackage[T1]{fontenc}
\usepackage{wrapfig}
\usepackage{placeins}
\usepackage{xspace}

\usepackage[capitalise,noabbrev,nameinlink]{cleveref}
\crefformat{equation}{#2Equation~#1#3}
\crefmultiformat{equation}{Equations~#2#1#3}{ and~#2#1#3}{, #2#1#3}{ and~#2#1#3}

\usepackage{enumitem}
\usepackage[english]{babel}
\usepackage[normalem]{ulem}

\setcitestyle{square}

\newcommand{\oursname}{EyeNeRF\xspace}
\newcommand{\nerfname}{NeRF-SHL\xspace}
\definecolor{gray}{gray}{0.4}

\newcommand{\hiddencomment}[1]{\iftrue {\color{gray}#1}\else {}\fi}  
\newcommand{\suppcomment}[1]{\iftrue {\color{red}[move to supp:] #1}\else {}\fi}  
\newcommand{\editcomment}[1]{\iftrue {\color{green}[edit:] #1}\else {}\fi}

\newcommand{\x}[0]{\mathbf{x}}
\newcommand{\wi}[0]{\omega_i}
\newcommand{\wo}[0]{\omega_o}

\begin{document}
	
\title{\oursname: A Hybrid Representation for Photorealistic Synthesis, Animation and Relighting of Human Eyes}

\author{Gengyan Li}
\orcid{0000-0002-1427-7612}
\email{gengyan.li@inf.ethz.ch}
\affiliation{%
	\institution{ETH Zurich and Google Inc.}
	\country{Switzerland}
}
\author{Abhimitra Meka}
\orcid{0000-0001-7906-4004}
\email{abhim@google.com}
\affiliation{%
	\institution{Google Inc.}
	\country{USA}
}
\author{Franziska Mueller}
\orcid{0000-0003-2036-9238}
\email{franziskamu@google.com}
\affiliation{%
	\institution{Google Inc.}
	\country{Switzerland}
}
\author{Marcel C. Buehler}
\orcid{0000-0001-8104-9313}
\email{buehlmar@ethz.ch}
\author{Otmar Hilliges}
\orcid{0000-0002-5068-3474}
\email{otmar.hilliges@inf.ethz.ch}
\affiliation{%
	\institution{ETH Zurich}
	\country{Switzerland}
}
\author{Thabo Beeler}
\orcid{0000-0002-8077-1205}
\email{tbeeler@google.com}
\affiliation{%
	\institution{Google Inc.}
	\country{Switzerland}
}

\renewcommand\shortauthors{Li et al.}

\begin{abstract}
	A unique challenge in creating high-quality animatable and relightable 3D avatars of real people is modeling human eyes, particularly in conjunction with the surrounding periocular face region. The challenge of synthesizing eyes is multifold as it requires 1) appropriate representations for the various components of the eye and the periocular region for coherent viewpoint synthesis, capable of representing diffuse, refractive and highly reflective surfaces, 2) disentangling skin and eye appearance from environmental illumination such that it may be rendered under novel lighting conditions, and 3) capturing eyeball motion and the deformation of the surrounding skin to enable re-gazing. 

These challenges have traditionally necessitated the use of expensive and cumbersome capture setups to obtain high-quality results, and even then, modeling of the full eye region holistically has remained elusive. We present a novel geometry and appearance representation that enables high-fidelity capture and photorealistic \textit{animation, view synthesis and relighting of the eye region} using only a sparse set of lights and cameras. Our hybrid representation combines an explicit parametric surface model for the eyeball surface with implicit deformable volumetric representations for the periocular region and the interior of the eye. This novel hybrid model has been designed specifically to address the various parts of that exceptionally challenging facial area - the explicit eyeball surface allows modeling refraction and high frequency specular reflection at the cornea, whereas the implicit representation is well suited to model lower frequency skin reflection via spherical harmonics and can represent non-surface structures such as hair (i.e. eyebrows) or highly diffuse volumetric bodies (i.e. sclera), both of which are a challenge for explicit surface models. Tightly integrating the two representations in a joint framework allows controlled photoreal image synthesis and joint optimization of both the geometry parameters of the eyeball and the implicit neural network in continuous 3D space. We show that for high-resolution close-ups of the human eye, our model can synthesize high-fidelity animated gaze from novel views under unseen illumination conditions, allowing to generate visually rich eye imagery.
\end{abstract}

\begin{teaserfigure}
    \centering
	\includegraphics[width=\textwidth]{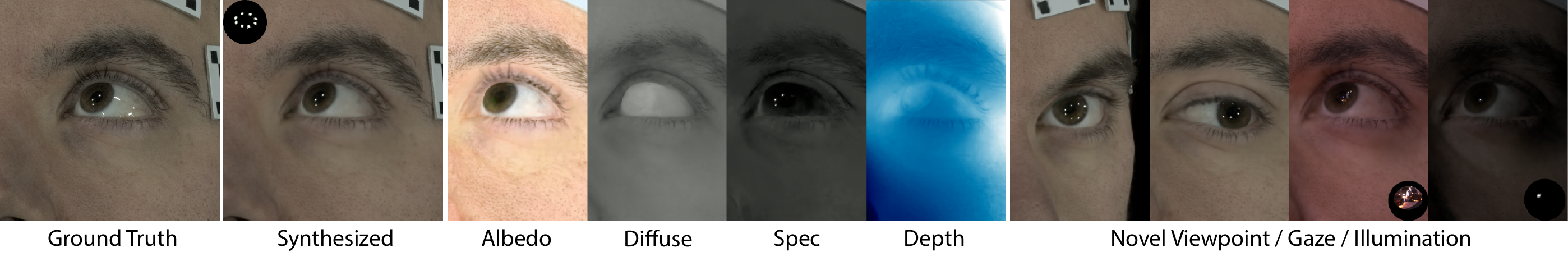}
	\caption
	{
		We introduce \oursname, a novel hybrid representation for photorealistic modeling and synthesis of human eyes. Our technique combines an explicit mesh based representation for the eyeball surface with an implicit neural representation for the periocular region (skin, eyelashes, eyelids etc.) and allows to synthesize photoreal images with novel gaze from a novel viewpoint and under novel illumination.
	}
	\label{fig:teaser}
\end{teaserfigure}

\maketitle


\section{Introduction}
\label{sec:introduction}

``Eyes are windows to the soul'' -- while an old adage, it captures what many media creators believe is an essential medium of human expression. There is no dearth of close up shots of actor's eyes on screen - from expressing horror in Hitchcock's Psycho (1960) to showing excitement in Scorsese's Goodfellas (1990) - such shots are a favorite tool in a director's toolbox. Recent research backs the art; \citet{doi:10.1177/0956797616687364} show that eyes provide not only important social cues, but are overwhelmingly interpreted as diagnostic of the subject's emotional state even in presence of competing signals from the lower face. Thus, it is only natural that as the graphics community embarks on another wave of photorealistic 3D avatar technologies, sufficient attention is paid to the development of methods and systems that allow for fine-grained photorealistic control over human eye imagery.

\begin{figure}
    \centering
    \includegraphics[width=\linewidth]{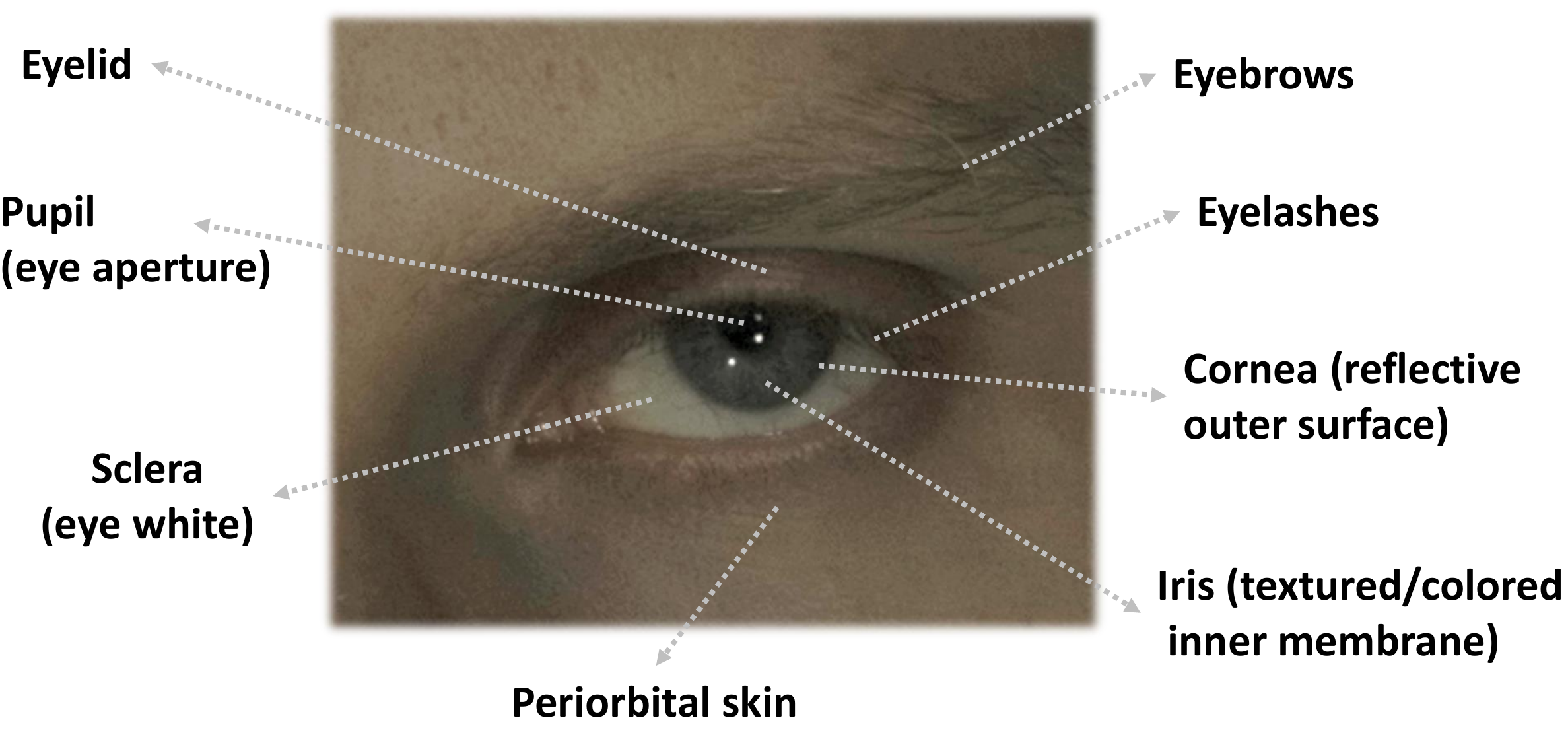}
    \caption{The eyeball and the periocular region exhibit very different geometry, motion and appearance characteristics. The eyeball is rigid, rotating and reflective/refractive, whereas the surrounding region is non-rigid, smoothly deforming and light-scattering. Hence, we use different representations to model each of these parts.}
    \label{fig:eyeanatomy}
\end{figure}

Modeling the eye region is challenging due to its complex anatomy (see Fig. \ref{fig:eyeanatomy}). The eyeball is a nearly smooth and rigid spheroid, which experiences negligible deformation. Its outer surface consists of a thin clear layer which is highly reflective. The corneal bulge is the protruding center that allows light to enter the eye. The cornea refracts light rays into the eyeball, which are further concentrated by the iris sphincter into the pupil. On the other hand, the surrounding region of the eye consists of multiple non-rigid and deforming materials like skin, eyelashes and eyebrows with fine geometry that exhibit light scattering effects. The motion of the eyeball is controlled by extraocular muscles that allow for fast rotation, while the surrounding skin exhibits smooth non-linear deformation. 

Modern eye reconstruction methods such as \citet{practicaleyerig, Berard16Eyes, 10.1145/2661229.2661285} achieve impressive results; they synthesize close to photorealistic imagery of the eye, depending on the amount and quality of data captured. Such methods however generally do not reconstruct the periocular region including the eyelids, eyelashes, eyebrows and periorbital skin, which are key to capturing eye expressions such as squinting, drooping, widening etc. An important step towards holistic modeling of the eye region is the work of \citet{schwartz2020eyes}, which proposes a technique to drive eye gaze in mesh based 3D avatars for VR communication. They achieve an impressive level of control over animation of the eye gaze and consistent deformation of the surrounding skin. While their application warrants real-time performance, it suffers from other drawbacks that come with mesh based 3D reconstruction techniques -- it cannot model fine structures of eyebrows and lashes and it requires a dense multi-view capture setup to reconstruct the mesh in a pre-processing step. Their method does not disentangle the complex reflectance of the eye and skin from the scene illumination, which makes it unsuitable for the applications of high-quality image synthesis under desired lighting environments. 

In the more general setting, \citet{park2021nerfies} introduce Deformable Neural Radiance Fields, also known as Nerfies, a generalization of the popular Neural Radiance Fields (NeRF) \cite{mildenhall2020nerf}, which can be used to reconstruct and synthesize the entire face and upper body through casual capture using a single hand-held moving camera. This and other similar techniques including \cite{park2021hypernerf, tretschk2021nonrigid, pumarola2021d} use volumetric rendering in continuous 3D space using a multi-layer perceptron to model the canonical shape of an object and learn a warp field on top of it to model dynamic deformations in the video sequence on a per-frame basis. While these techniques excel at modeling both dynamic deformations of skin and thin structures such as hair due to the underlying volumetric representation, they are not animatable and do not provide a solution to modeling surface reflectance and high-frequency light-transport effects such as corneal light reflection and refraction associated with the eye, which prohibits relighting. Works such as \citet{bi2020deep, nerv2021, physg2020, 10.1145/3478513.3480496} have extended volumetric rendering models to further disentangle reflectance and scene lighting, initially using single point light sources co-located with the camera to illuminate the scene and later generalizing to unconstrained lighting, but only under rigid and static settings.

Inspired by these techniques, we propose a novel hybrid representation that combines the best of mesh based and volumetric reconstruction to achieve animatable synthesis of the eye region under desired environmental lighting. We use a light-weight capture system consisting of a small number of static cameras and lights along with a single hand-held camera with a co-located light source. We model the eyeball surface as an explicit mesh and the canonical shape of the periocular skin region and the interior eye volume using an implicit volumetric representation. The eyeball mesh is used to explicitly compute specular reflections of light rays as well as refraction of the camera rays at the cornea surface. The deformations of the surrounding skin and hair is computed using a learnt warp field over the canonical volume. In order to achieve relightability, we learn the underlying reflectance represented by spherical harmonics coefficients. The outgoing radiance is then computed as a product of the reflectance and environmental illumination in the 3D frequency domain. We jointly optimize for the shape and pose of the eyeball mesh and the density and reflectance in the implicit volume, supervised to minimize the photometric loss between the modeled outgoing radiance and pixel values in the captured video.

We show that our method is able to successfully reconstruct the canonical geometry of the eye region and model it's appearance by accurately disentangling shading from diffuse and specular albedos (see Fig.~\ref{fig:teaser}). This enables photorealistic view synthesis and relighting by recomputing the shading under novel environmental illumination. Moreover, by interpolating between the learnt warp field of the captured frames and rotating the explicit eyeball mesh, we achieve fine-grained control over the subject's gaze. We demonstrate these capabilities of our method on several subjects with variation in face and eye appearance. We summarize our contributions as follows:

\begin{enumerate}
    \item We propose a hybrid mesh + implicit volumetric representation that allows for modeling of complex reflectance and fine scale geometry of the eye region.
    \item We design a capture system using only off-the-shelf hardware that allows capturing data to disentangle appearance from scene illumination to achieve high-frequency relighting.
    \item We demonstrate exciting animated and relit results on several real subjects with varying facial and ocular characteristics.
\end{enumerate}

\section{Related Work}
\label{sec:relatedwork}

Automatic reconstruction of human eye and face involves complex modeling of geometry, appearance and deformation. Our work builds on several lines of work that we discuss in greater detail here.

\paragraph{Eye Modeling and Synthesis}

Photorealistic control of real eye imagery has been important to various application domains. Image-based  techniques such as \cite{Buehler2019ICCVW, Zheng2020NeurIPS, he2019gazeredirection, Kaur_2021_WACV, Kaur_2020_WACV, 7299098,ganin2016deepwarp,DBLP:conf/iclr/LeeKS18,8099724} have used eye keypoints, gaze direction or underlying segmentation masks as control signals to synthesize eye images for a given subject. Such works are primarily focused on gaze redirection or gaze correction application for portrait images and videos. A complementary line of work \cite{7122896,8578151,zhang15_cvpr,9022317,6909631} explores appearance based gaze estimation using analysis by synthesis techniques. Real-time control of gaze is also crucial to avatar based communication in AR/VR settings. \citet{thies16FaceVR, thies2018headon} propose facial reenactment systems using a head-mounted display (HMD) that performs gaze estimation using IR cameras inside the device and photorealistically resynthesizes the subject's face from an external camera with desired gaze direction. \citet{schwartz2020eyes} extend this application by fully modeling the eyeball using a parametric eye model. They also model view-dependent texture and periocular deformation using a neural rendering network. While these technique excel at controlling gaze, they do not model the complex reflectance characteristics of the eye and hence cannot resynthesize it under novel lighting. 

\citet{10.1145/2661229.2661285} propose a capture system to model subject-specific eyeball characteristics such as the white sclera, the transparent cornea, and the non-rigidly deforming colored iris. Using multi-view close-up images of the eye, they reconstruct the geometry, appearance and deformation of individual parts of the eye to generate a high-quality graphical model for computer graphics applications. \citet{Berard16Eyes} extend this work by leveraging a database of eye images to generate a prior to constrain the eyeball shape and iris texture and deformation. This enables a lightweight technique to generate graphics models of eyes even from single in-the-wild close up photographs. \citet{practicaleyerig} focus on rigging a parametric eye model by reconstructing accurate eye poses from a multi-view capture. They estimate subject-specific geometry properties such as eyeball shape, rotation center, inter-ocular distance and visual axis by optimizing the reprojection error of manually annotated 2D eye keypoints and contours. These techniques aim at modeling the appearance and geometry of the eyeball in great detail, but do not model the fine scale structure and deformations of the periocular face region.

\paragraph{Volumetric Deformable Surfaces}

Neural Volumes \cite{Lombardi:2019} and NeRF \cite{mildenhall2020nerf} introduced implicit neural network based volumetric reconstruction from multi-view imagery that enables photorealistic novel view synthesis of scenes. Such a representation is particularly well suited for capturing view-dependent appearance, such as on skin, and volumetric effects, including hair, and hence is a suitable representation for capturing digital versions of real people. But in order to realistically animate such a model, it is important to also learn the modes of deformation of skin and hair.  \cite{Gafni_2021_CVPR} densely tracked a face in a monocular video sequence using a parametric face model and reprojected it into 3D space to learn an implicit canonical volume. This allowed them to control face pose and also perform dynamic animations with it by modifying the face model parameters at test time. But this method relies on a differential face model to model the deformation. 

\citet{park2021nerfies} introduced an extension to NeRF, called Nerfies, that learns a canonical volume of a deforming object such as the upper torso of a person, and also generates a learnt per-frame deformation field that warps the volume in order to synthesize the corresponding ground-truth image. \cite{tretschk2021nonrigid, xian2021space, li2020neural,pumarola2021d} concurrently proposed methods with similar underlying principles. These methods showed that it is possible to not only extract neutral volume density of a non-rigid scene, but also learn the modes of deformation of such a volume without any priors, using only monocular video capture from a freely moving camera.

\paragraph{Reflectance Modeling and Relighting}

Methods such as \cite{9010718, 10.1145/3095816}  have achieved environmental relighting of portrait images from casual monocular capture, but produce only diffuse lighting effects and hence do not achieve photorealism. Amongst image-based techniques, \citet{debevec_acquiring_2000} was the first to achieve accurate high-frequency relighting of a static face by acquiring the full reflectance field using a Light Stage, a dome containing densely distributed calibrated lights and cameras. More recently, Light Stage data has been used to train neural rendering techniques to relight in-the-wild images \cite{Pandey:2021, mbr_frf, mallikarjun2021photoapp}, dynamic sequences \cite{Meka:2019}, animatable avatars \cite{Bi-2021-128861} and  denser portrait reflectance fields \cite{sun2020light}. Differentiable ray-tracers have also been used to inverse render portrait images to disentangle reflectance and scene lighting \cite{dib2021high,physg2020}.

Amongst implicit volumetric reconstruction techniques, \citet{bi2020deep} were one of the first to estimate higher-order surface reflectance and visibility in the scene. They use unstructured flash images from a camera with a collocated point light source to learn volume density and reflectance (parameterized as Disney BRDF parameters of diffuse albedo and specular roughness), enabling them to relight the scene under any point light source. \citet{sun2021nelf} train their model on synthetic data to estimate the light transport field in 3D from multi-view portrait images. Their method achieves convincing reflectance decomposition and relighting, but is limited by the photorealism of the synthetic training data. \citet{10.1145/3478513.3480496, zhang2021ners,boss2021nerd,nerv2021} extend volumetric scene representation to further factorize the radiance field in dense lighting and reflectance and other components such as visiblity and indirect illumination, enabling full environmental relighting of the scene. These techniques are constrained to rigid and static settings.

Our method takes inspiration from explicit eyeball modeling methods of \citet{practicaleyerig, Berard16Eyes, 10.1145/2661229.2661285} for the eyeball surface, the deformable volumetric reconstruction of \citet{park2021nerfies} for the periocular region and the volumetric relighting technique of \citet{bi2020deep} to disentangle reflectance from environmental lighting using a sparse multi-view capture setup including a single handheld freely-moving camera with a co-located light source, and the lighting visibility model from \cite{nerv2021} which avoids expensive secondary reflection rays by approximating self occlusion using a neural network.

\section{Hybrid Model}
\label{sec:method}
The eye region is comprised of components with vastly different visual properties. The surface of the eye is so specular that it mirrors the environment \cite{Nishino2006CornealImaging} overlaid over the underlying iris, which is heavily distorted through the optical refraction at the corneal surface, especially for side-views. The white sclera is highly scattering, exhibiting veins at different depths inside. The eye is embedded in the periocular region, further adding to the challenge as it combines highly deforming skin and hair from lashes and brows. To address this large diversity, we propose a novel hybrid model that combines the strengths of explicit and implicit representations. In this section we describe the individual parts of the model and how they fit together.

\subsection{Explicit Eyeball Surface Model}
\label{sec:expliciteyemodel}

\begingroup

To model the highly reflective and refractive eyeball surface we represent it with an explicit parametric shape model--concretely we employ a variant of the LeGrand eye model~\cite{legrand1957}, which consists of two overlapping spheres, but other parametric models could be used as well. 
The model is fully parameterized by 3 parameters: iris radius $b$, which specifies the width of the intersecting circle 
\setlength{\columnsep}{2pt}%
\setlength\intextsep{0pt}%
\begin{wrapfigure}[10]{r}{0.38\columnwidth}
\begin{center}
    \includegraphics[width=0.38\columnwidth,trim={0cm 0cm 0cm 0.4cm}, clip]{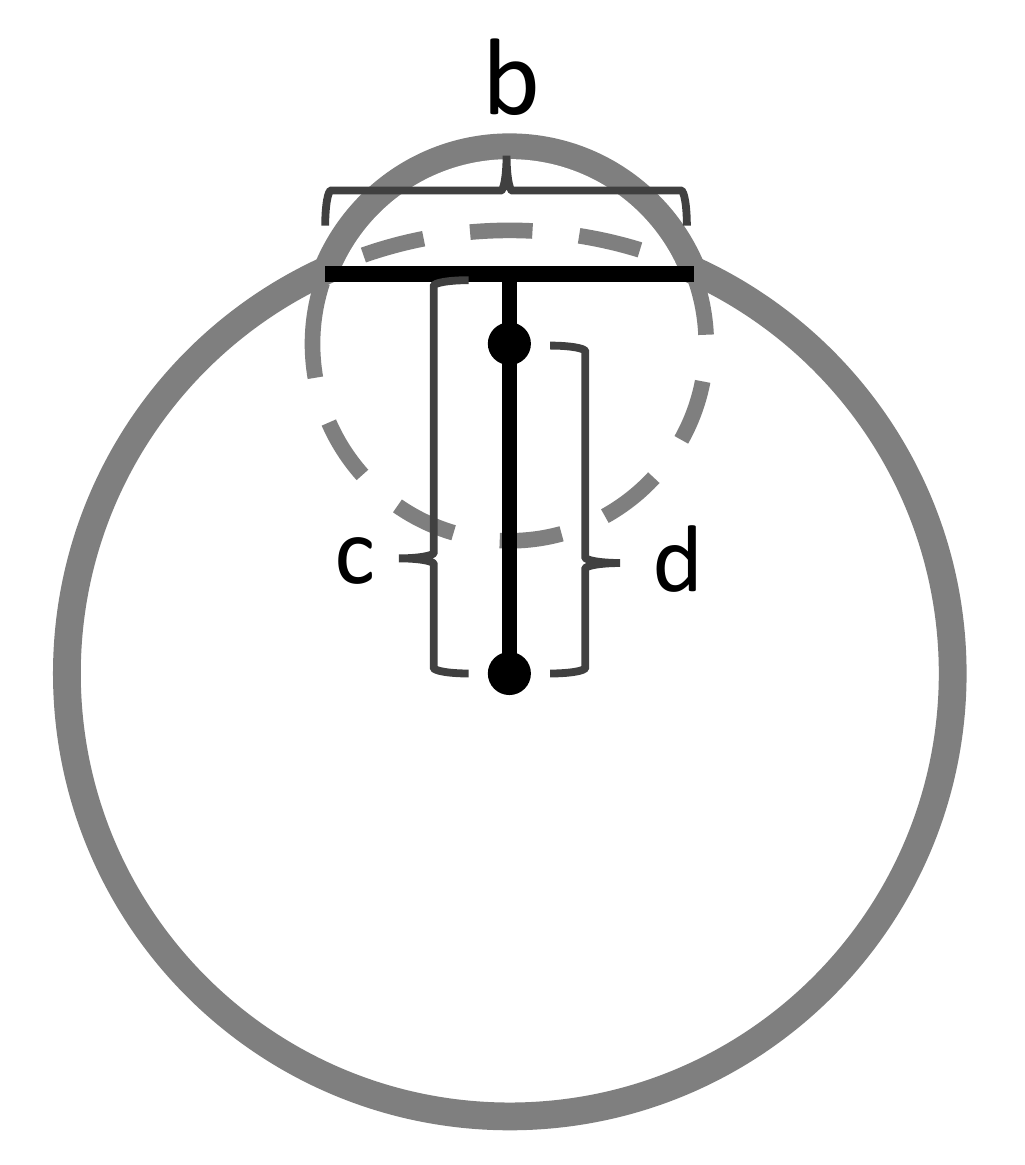}
\end{center}
\end{wrapfigure}%
of the eyeball and cornea spheres, the iris offset $c$ which specifies the distance of the aforementioned circle from the eyeball center, and the cornea offset $d$, which specifies the relative distance of the two sphere centers. 
These values can then be used to derive the main eyeball radius as well as the cornea radius.
Akin to \citet{schwartz2020eyes} we smoothly blend the eyeball and corneal sphere at the transition (limbus). 
This blending is controlled by two additional learnable parameters, determining where the transitions on the eyeball and corneal spheres start.
The model is discretized as a triangular mesh with 10242 vertices, and enriched with per-vertex displacements, which enable the surface to represent shapes that lie outside the model's subspace. We then compute the shading normals at each vertex (used for refraction and reflection) by interpolating between the neighboring face normals weighted by their vertex angle.
For the index of refraction (IOR), we use the value of 1.4, which is a reasonable value for the human cornea~\cite{defreitas2013vivo, patel1995refractive}.
While we assume the eyeball surface to remain static for a subject, its pose changes as a function of gaze. Though eye gaze is often modelled as a 2-DoF rotation only, it is actually more complex than that \cite{practicaleyerig} and hence we model its motion by a 6-DoF transformation per frame, encoded in axis angle representation, with translation being applied after rotation.

\endgroup

\begin{figure*}
    \centering
    \includegraphics[width=\textwidth,page=1,trim={0 6.5cm 0 0},clip]{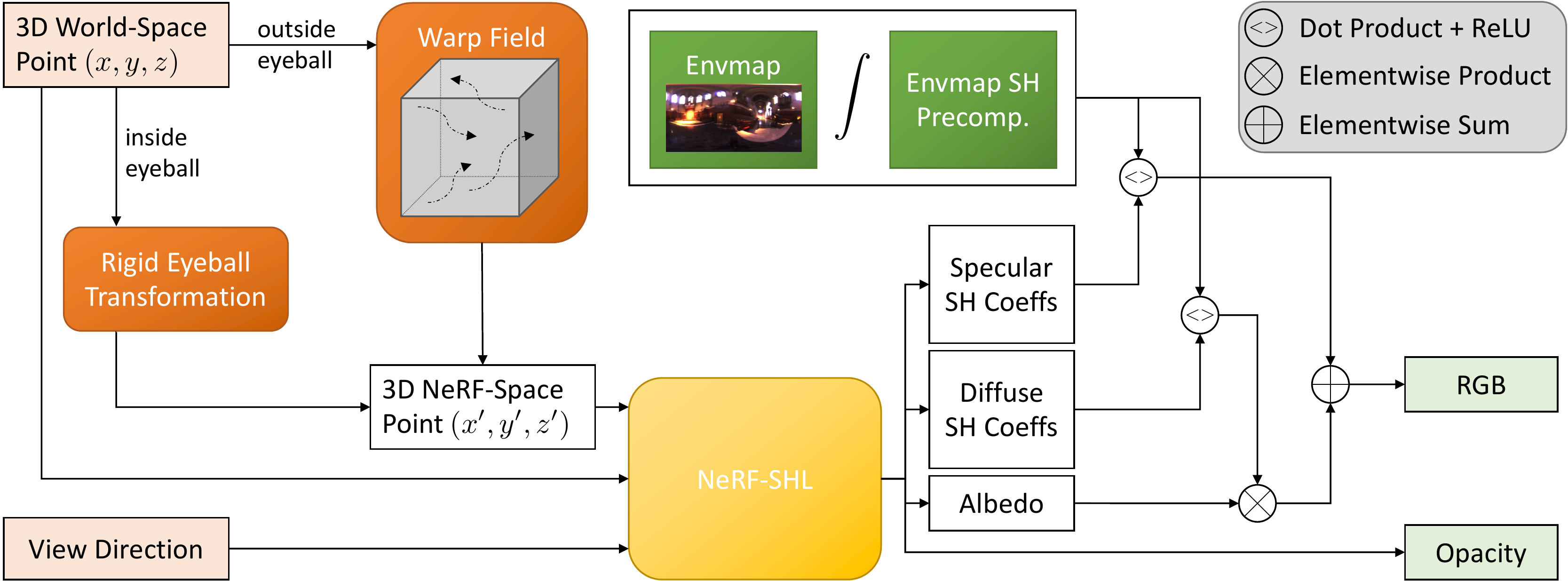}
    \caption{
    \textbf{Hybrid Model Evaluation.} For every 3D point in world space and a view direction, our hybrid model is trained to output the corresponding RGB color and opacity value.
    We first transform the 3D world-space point to the canonical NeRF space by using the learned per-frame rigid eyeball transformation or warp field, depending on whether the point lies inside the eyeball.
    Next, we evaluate our \nerfname to obtain the opacity, the albedo, and specular and diffuse spherical harmonics (SH) coefficients.
    The SH coefficients are multiplied with the pre-computed SH representation of the environment map and composited with the albedo to obtain the final RGB color for the 3D point.
    }
    \label{fig:overview}
\end{figure*}

\subsection{Implicit Eye Interior and Periocular Model}
The highly scattering sclera, the volumetric iris, the transition between eyelid and eye, and especially hair fibers present a formidable challenge for explicit models, and so we argue that an implicit representation is better suited to represent the periocular region and to seamlessly integrate with the eye. Since the skin deforms during acquisition we base our representation on Nerfies~\cite{park2021nerfies}, which employs warp fields to transform frames into a common canonical space, where an MLP network encodes opacity and appearance values as proposed by the seminal NeRF paper \cite{mildenhall2020nerf}. The warp field is defined by a secondary MLP which predicts a rotation quaternion and translation vector. We combine the smooth warp field proposed by Nerfies with the rigid transformation from the eyeball surface to explicitly transport rays to a canonical eye volume once they intersect the eyeball surface, where the NeRF encodes the interior.

\subsubsection{\nerfname}
\label{sec:nerf_nn}
\begin{figure}
    \centering
    \includegraphics[width=\columnwidth,page=3,trim={0 4.2cm 8cm 0},clip]{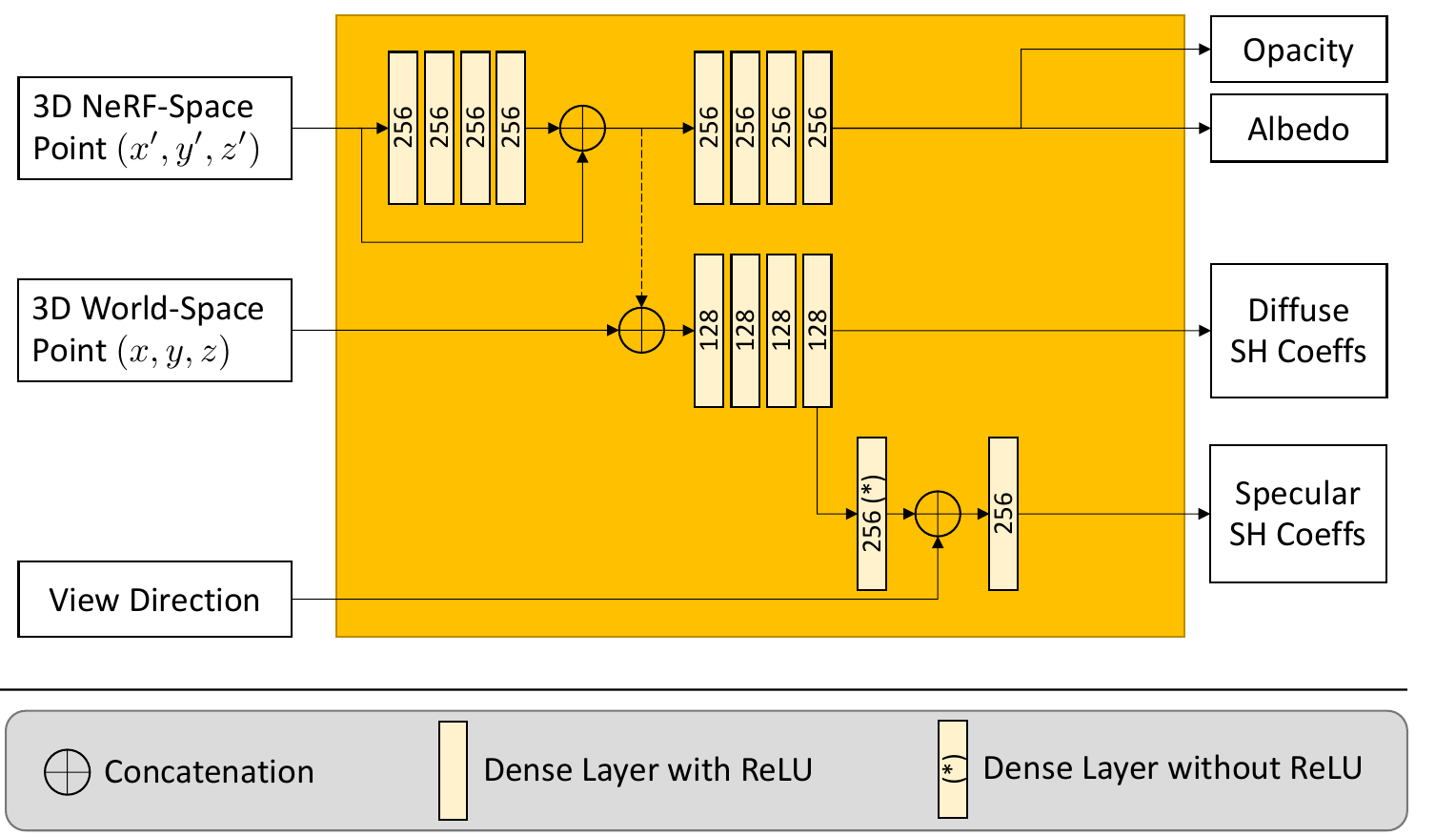}
    \caption{The architecture for \nerfname can be divided intro three branches. The first branch predicts opacity and albedo from the 3D point in canonical NeRF space. For the second branch, we additionally feed the 3D world-space point as input to better model shadowing. Lastly, we add the view direction input for the branch that predicts specular SH coefficients.}
    \label{fig:architecture}
\end{figure}

To enable relighting we propose \nerfname (NeRF with Spherical Harmonics Lighting), which extends the traditional NeRF network as proposed by \citet{mildenhall2020nerf} to additionally predict spherical harmonics coefficients alongside opacity and albedo. 
These coefficients are used to predict the exiting radiance given an environment map and implicitly approximate various light-transport effects, including reflectance, subsurface scattering, occlusion, and indirect illumination. To better model the combination of diffuse and specular reflectance, we predict two sets of SH coefficients, using 5th order SH for the diffuse and 8th order SH for the specular reflectance. As diffuse reflectance is constant with regards to the outgoing light direction, only the specular SH coefficients are conditioned on the view direction.
The network takes as input a 3D World-Space point and its corresponding 3D NeRF-Space point alongside their positional encodings as well as the view direction and outputs diffuse and specular spherical harmonics coefficients for the query point (Fig.~\ref{fig:overview}). These are integrated with the environment illumination and combined to produce the RGB value.
As in the original NeRF network, we constrain the opacity to be positive using a relu activation, and we apply a sigmoid to the albedo, similar to how the RGB is constrained in the original network. Additionally, we apply a softplus activation function to the 0th degree spherical harmonics function to force it to be positive, as that particular spherical harmonic corresponds to the uniform function which needs to be positive for any physically correct light transport function. A schematic of the network architecture is shown in Fig.~\ref{fig:architecture}.

\subsection{Illumination Model}
\label{sec:illumination_model}
Environmental illumination is represented as a lat-long environment map $E(\wi)$, which determines the amount of radiance entering the scene from a direction $\wi$. While relighting the explicit eyeball surface can consume the environment map directly, it needs to be converted to a spherical harmonics representation for relighting the implicit parts of the model to be compatible with our network architecture as introduced in Sec.~\ref{sec:nerf_nn}.

\paragraph{Spherical Harmonics Precomputation.}
\label{sec:env_map}

We start with the standard light transport equation (without emission), at any given point $\x$ and outgoing light direction (or camera direction) $\wo$:
\begin{equation}
L_o(\x, \wo) = \int_{\Omega} f(\x, \wo, \wi) L_i(\x, \wi) d \wi \: .
\end{equation}
In our model, $L_i$ includes all incident light at position $\x$, i.e. light from both direct and indirect light as well as incident illumination from subsurface scattering, and hence $f$ approximates the full light transport at $\x$ for the entire sphere $\Omega$. We can therefore reformulate this equation to instead determine the amount of light transported from the environment map $E$ via a point $\x$ towards $\wo$ as
\begin{equation}
    L_o(\x, \wo) = \int_{\Omega} f_{\text{tot}}(\x, \wo, \wi) E(\wi) d \wi \: .
\end{equation}
While intractable with traditional computer graphics methods, it is possible to approximate $f_{\text{tot}}$ using machine learning techniques.
We can approximate $f_{\text{tot}}$ using the spherical harmonics basis functions $Y_{lm}$ and their corresponding coefficients $c_{lm}$
\begin{equation}
    f_{\text{tot}(\x, \wo, \wi)} \approx \sum_{l=0}^{\text{order}} \sum_{m=-l}^{l}c_{lm}(\x, \wo)Y_{lm}(\wi) \: ,
\end{equation}
resulting in our approximation
\begin{equation}
    L_o(\x, \wo) \approx \int_{\Omega} \sum_{l=0}^{\text{order}} \sum_{m=-l}^{l}c_{lm}(\x, \wo)Y_{lm}(\wi) E(\wi) d \wi \:.
    \label{eq:approx_1}
\end{equation}
By using associativity of sums and integrals as well as distributivity of sums and products, we can reorder Eq.~\ref{eq:approx_1} such that the integral can be precomputed independently of the coefficients
\begin{equation}
   L_o(\x, \wo) \approx \sum_{l=0}^{\text{order}} \sum_{m=-l}^{l}c_{lm}(x, \wo)\left(\int_{\Omega}Y_{lm}(\wi) E(\wi) d \wi\right) \: . 
\end{equation}
Most notably, the integral is now independent of the conditioning variables, $\x$ and $\wo$, which allows to compute the integral once for each spherical harmonics order and degree for a given environment map, making model training in Sec.~\ref{sec:training} tractable.
As described in Sec.~\ref{sec:data} we capture our subjects under a mixture of static environment illumination and a moving point light, and precompute a spherical harmonics representation for both. We rotate these representations appropriately to compensate for head-rotation, which can be done efficiently as outlined in Appendix~\ref{app:SH_rotation}. Lastly, the SH coefficients of the moving light are scaled as a function of the distance to the subject to account for squared intensity falloff and the two sets of coefficients are summed to allow relighting of our implicit model.

\section{Model Evaluation}
The presented model can be evaluated for a desired gaze direction and rendered from novel viewpoints under novel illumination.

\subsection{Gaze Animation}
Trained on a diverse set of discrete gaze directions, \oursname allows continuous re-animation of the eye and surrounding region by interpolating the deformation fields and eyeball poses of the discrete training data.
Given a novel gaze direction $\gamma$ the first step is to identify three gaze directions $\hat{\gamma}_i$ which form a convex hull that contains the target gaze direction. For this, we project all training gaze directions onto the unit sphere by applying their respective rigid eyeball transforms to the unit vector $(0,0,1)^\top$ and triangulate those points using the Ball-pivoting Algorithm~\cite{fausto1999ball,Zhou2018open3d} to obtain a discrete mesh. 
We compute the intersection of the target gaze direction $\gamma$ with this mesh. The vertices of the intersected triangle are the desired discrete gazes $\hat{\gamma}_i$ and the barycentric weights of the intersection point serve as interpolation weights which are then used to blend the warp fields of the three sample gaze directions by linearly interpolating between the warped points. This method limits novel gazes to be within the gaze distribution from the training set, which is reasonable since we capture the full range of motion of the eye during acquisition.

For the eyeball transformation, we consider the translation and rotation components separately. The translation is interpolated using barycentric weights as done for the warp field.
The rotation on the other hand is more challenging, since the basic spherical linear interpolation (slerp) leads to unnatural eye motion. Instead, we independently rotate the eyeball for each sample gaze to the target gaze, and slerp the resulting poses sequentially. We found that this yields satisfactory eye motion, but more advanced eye rigging methods, such as the Listings model used in \cite{practicaleyerig}, could also be integrated with the proposed hybrid model.

\subsection{Rendering}
\label{sec:raymarching}

\begin{figure}
    \centering
    \includegraphics[width=\columnwidth]{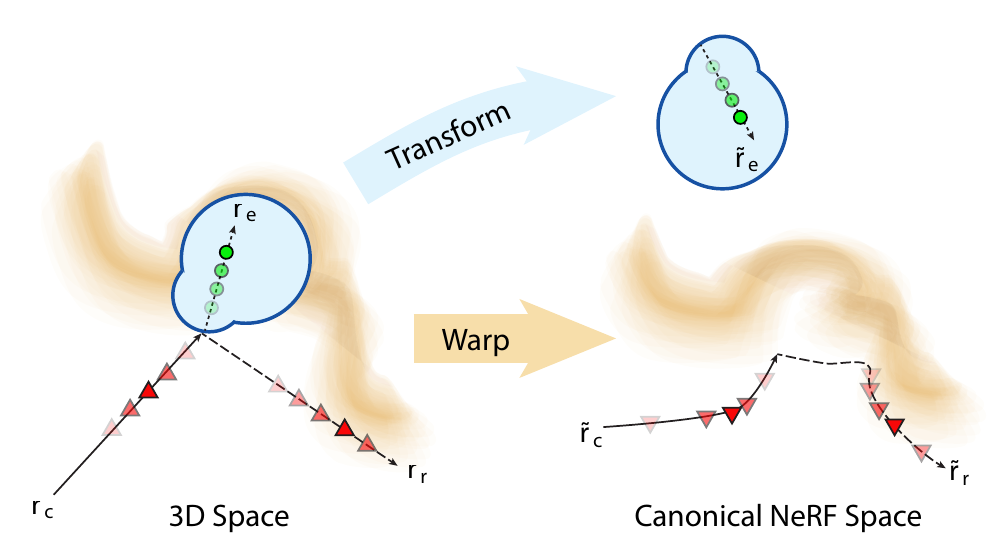}
    \caption{\oursname uses raytracing to compute reflection $r_r$ and refraction $r_e$ rays by intersecting with the explicit eyeball surface. Those rays are then used to raymarch the implicit representation. Points are sampled in 3D Space, and then transformed to the canonical NeRF Space. Points sampled from the refraction ray $r_e$ are transformed rigidly by the inverse estimated eyeball pose (green points) where points sampled from the other rays are warped non-rigidly by the learned warp field (red triangles). }
    \label{fig:raymarching}
\end{figure}

Once the interpolated warp field and rigid transformation have been computed, the goal is to render an image for a given camera and environmental illumination. Fig.~\ref{fig:raymarching} provides a schematic of the approach presented below.

\subsubsection{Ray Computation and Intersection}
For each pixel in the image we compute the camera ray $r_c$ and trace it through the scene within appropriate clipping planes. We start by testing if the ray intersects with the explicit eyeball surface using Trimesh \cite{trimesh}, a raytracer implemented using Embree \cite{wald2014embree}. In case there is an intersection, we calculate appropriate reflection and refraction rays at the corneal surface using Snell's law, splitting the original ray into three parts: The pre-intersect ray $r_c$, the refracted ray $r_e$, and the reflected ray $r_r$. If there is no intersection, we only need to consider the pre-intersect ray $r_c$. 

\subsubsection{Point Sampling and Transformation}
\label{sec:sampling}
As depicted in Fig.~\ref{fig:overview} and Fig.~\ref{fig:raymarching}, we sample points along the three rays and transform them to the canonical NeRF volume. Following \cite{mildenhall2020nerf} we employ a combination of equidistant and importance sampling in 3D world-space to determine the sample points.
For each sample point of the pre-intersect and reflected rays, we evaluate the warp field neural network and warp the points accordingly, akin to \citet{park2021nerfies}, leading to sample points along the distorted rays $\tilde{r}_c$ and $\tilde{r}_r$ in the canonical 3D NeRF-space (red triangles in Fig.~\ref{fig:raymarching}).
For each sample point on the refracted ray, we apply the inverse of the rigid eyeball transform leading to sample points in the canonical 3D NeRF-space of the eye (green circles in Fig.~\ref{fig:raymarching}). 

\subsubsection{Point Shading}
Next, we calculate the contribution of the illumination from the environment map for each sample point.
We then query \nerfname to obtain albedo and opacity, as well as specular and diffuse SH coefficients that determine the amount of light transported from the environment map at the queried volume point towards the queried camera ray direction. Incident illumination and transfer function are integrated by multiplying the SH coefficients with the precomputed SH environment coefficients, which is very efficient.
The final color value for the sample point is then obtained by multiplying the diffuse lighting with the albedo and adding the specular lighting.
For the end point of each ray, i.e.\:when it is leaving the captured volume, we add one sample point with infinite opacity and a color value. For reflection rays the color is retrieved from the environment map and for the other rays it is set to black. To account for ambiguities in scale when reconstructing the environment map using the mirror ball, we also learn a scale factor which is multiplied with the environment map radiance sample.

\subsubsection{Color Accumulation}
The previous step provided opacity and color for each of the $N_S$ sample points independently.
To obtain the contribution weight of each sample to the final color value, we employ traditional volume rendering techniques~\cite{kajiya1984ray}.
To this end, we calculate the accumulated transmittance for each sample point based on the opacity of all previous samples on the ray. 
In the traditional NeRF setting, the color value of a single ray $C(\mathbf{r})$ is computed using the following approximation of the continuous integral along the ray
\begin{equation}
   \alpha_i = e^{-\sigma_i\delta_i} \qquad T_t = \prod_{i=0}^{t-1} \alpha_i \qquad C(\mathbf{r}) = \sum_{t=0}^{N_S} T_t(1-\alpha_t) \mathbf{c}_t 
\label{eq:accumulation_nerf}
\end{equation}
where $\sigma_i$ and $\mathbf{c}_i$ are the predicted opacity and color for the $i$-th sample, respectively, and $\delta_i$ represents the distance between the $i$-th and $i+1$-th sample.
We refer the reader to \citet{mildenhall2020nerf} for more details.
In our implementation, we first compute the color value of the reflected ray separately, and then merge the refracted and pre-intersect ray using the Fresnel Equations~\cite{fresnel1868oeuvres} (assuming unpolarized light), by combining the radiance of the reflected ray with the last sample prior to the intersection using standard alpha compositing rules, effectively placing it behind that sample. We can then treat the resulting combined samples as one ray, resulting in the following system of equations
\begin{equation}
\begin{split}
    &\alpha_i' = e^{-\sigma_i'\delta_i'} \qquad T_t' = \prod_{i=0}^{t-1} \alpha_i' \qquad C(\mathbf{r}') = \sum_{t=0}^{N_S'} T_t'(1-\alpha_t') \mathbf{c}_t' \\
    &\alpha_{\text{comb}} = \alpha_k(1-f) \qquad c_{\text{comb}} = \frac{(1-\alpha_k) \mathbf{c}_k + f\alpha_k  C(\mathbf{r}')}{\alpha_{\text{comb}}} \\
    &\mathbf{c}_i'' = \begin{cases}
      \mathbf{c}_i & i \neq k \\
      \mathbf{c}_{\text{comb}} & i = k \\
\end{cases} 
\qquad
\alpha_i'' = \begin{cases}
      \alpha_i & i \neq k \\
      \alpha_{\text{comb}} & i = k \\
\end{cases} 
\\
    &\alpha_i = e^{-\sigma_i\delta_i} \qquad T_t = \prod_{i=0}^{t-1} \alpha_i \qquad C(\mathbf{r}) = \sum_{t=0}^{N_S} T_t(1-\alpha_t'') \mathbf{c}_t''
\end{split}
\label{eq:accumulation_eyenerf_refl}
\end{equation}
where $f$ is the Fresnel factor, $k$ is the index of the last sample prior to the intersection, $\alpha'$, etc. refer to values sampled along the reflected ray, and $c_i''$ refers to the samples of the combined ray (which are the same as the ones along the pre-intersect and refracted ray, other than the sample prior to the eyeball intersection.

Rays that do not intersect the eye model are computed using the original NeRF ray marching method (see Eq.~\ref{eq:accumulation_nerf}).


\section{Model Training}
\label{sec:data}

High quality synthesis requires high-quality data. While there are several publicly available eye image datasets, they are unfortunately not directly suited for our purposes of view, gaze, and illumination synthesis. The majority of these datasets are tailored for the task of gaze-tracking \cite{wu2020magiceyes, kim2019, wood2015_iccv, zhang15_cvpr, 10.1145/3130971, Fusek2018433, fuhl2021teyed} while others cater to different problems such as pupil detection \cite{10.1145/2857491.2857520}, eye closure detection \cite{SONG20142825} or eyelash segmentation \cite{10.1145/3478513.3480540}. While there are datasets that aim at modeling high-quality eyes and periocular region \cite{10.1145/2661229.2661285, practicaleyerig}, these are not suited for relighting purposes.

Hence we built our own capture system that provides sufficient signal for the task of gaze reanimation, view synthesis and relighting of the periocular region. We detail our hardware setup and the capture protocol below.

\subsection{Capture System}
\label{sec:capture-system}

\begin{figure}
    \centering
    \includegraphics[width=\columnwidth]{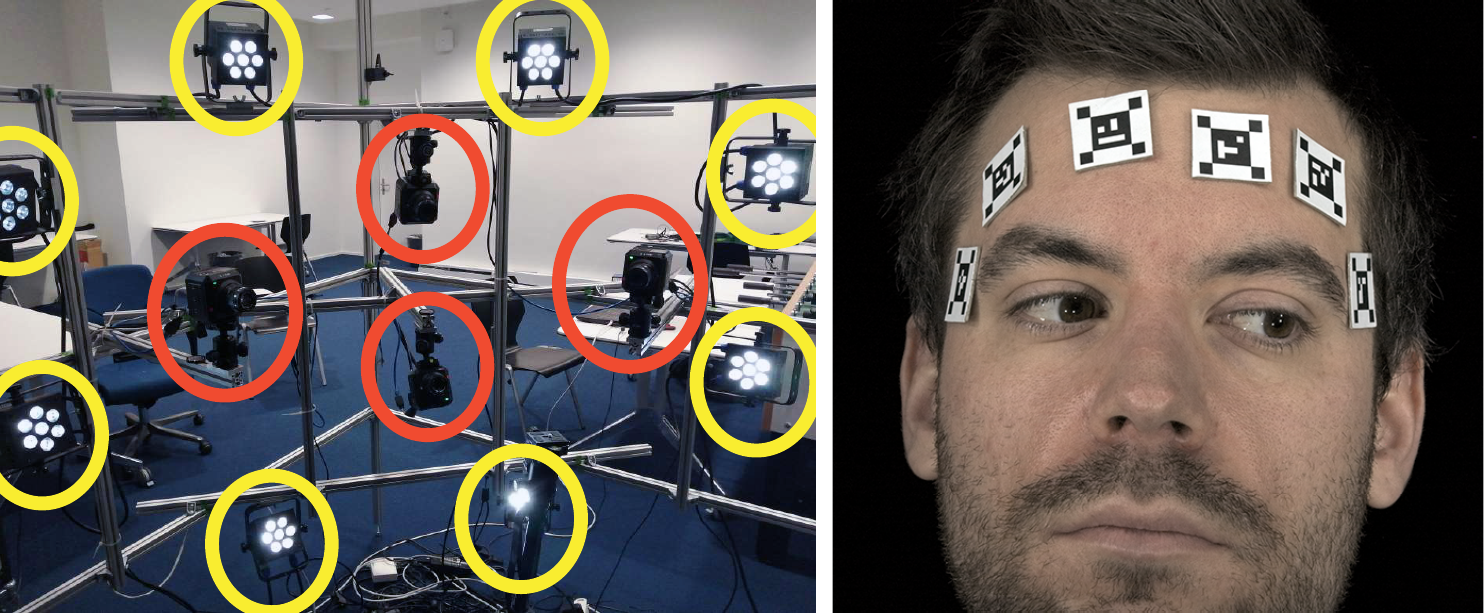}
    \caption{\textbf{left:} Our static setup consists of 4 high-quality cameras (red) arranged in a diamond-shape and surrounded by 8 illuminators (yellow). \textbf{right:} A set of AR markers is attached to the forehead of the subject to track head movement as well as relative camera motion. }
    \label{fig:setup}
\end{figure}

We aim to minimize the complexity of our hardware to make our solution cost and space effective. The subject sits on a chair in the center of the setup shown in Fig.~\ref{fig:setup}. Our multi-view setup consists of 4 high quality, hardware synchronized cameras (Z-Cam E2 4K) fixed in the frontal hemisphere of the subject, as well as a small, lower quality mobile GoPro\textsubscript{\textregistered} Hero9 camera which is moved freely by hand by the operator. The moving camera contains a co-located LED light, which is crucial for relighting, as we demonstrate in the next section. The subject is lit with 8 white LED point lights that are nearly uniformly located over a $100^{\circ}$ field-of-view (FOV) in front of the subject. The cameras span roughly $75^{\circ}$ horizontal FOV and $25^{\circ}$ vertical FOV in front of the subject, arranged in a diamond shape.

In order to account for the free head motion of the subject, we affix a set of small calibration markers on their forehead. These calibration markers are used to localize both the subject's head and the moving camera with respect to the static setup. While we assume that the eyeball is rigidly attached to the head, our method makes no such assumption for the periocular region which can experience strong deformations due to facial muscles and skin. 

\subsection{Capture Protocol}
\label{sec:capture-protocol}

We capture subjects under four different conditions.
In the first condition, the subject is instructed to follow the mobile camera with their gaze while keeping their head static and forward facing. The mobile camera is translated freely, while orienting it towards the subject's head, covering about $60^{\circ}$ horizontally and $30^{\circ}$ vertically. In this setting, we only use the 8 static lights. Since the eye gaze follows the mobile camera, we know the gaze direction and get a good multi-view coverage of the eye from the 4 static cameras.

In the other three conditions, the subject is instructed to keep their gaze focused on one of the four static cameras, switching their gaze between them when instructed. Instead of moving their gaze, they are instructed to rotate their head around, while attempting to keep one eye roughly stationary, to keep that eye in frame and at the same distance from the cameras. In order to more easily keep track of where that eye was, we place a tripod below the subject's head for reference (not as chin rest). In these settings we get good viewpoint coverage of the periocular region from the 4 cameras.

Each of these three conditions models a different illumination scenario; static lights, mobile light, and both. This provides a good mixture of frames where we have relatively flat lighting which is useful for reconstructing the geometry, and very high frequency lighting which can be used to learn shading and shadowing.

For the demonstrations in this paper, we capture the performances of 3 subjects with a variety of face shapes and eye colors.
\subsection{Data Preprocessing}
\label{sec:dataprocessing}
\subsubsection{Cameras}
Intrinsics for all cameras are calibrated using a checkerboard pattern. Extrinsics of the static cameras are recovered from the same calibration process. The extrinsics of the mobile camera are estimated from the marker tags on the subjects forehead using OpenCV. By also estimating the rigid transformation between the static cameras and the subject we can relate all cameras into the same world frame, registered to the subjects head.

\subsubsection{Environment Illumination}
We compute an HDR environment map from a series of images from a mirror sphere with varying exposure, similar to \cite{10.1145/344779.344855}. We capture an environment map for the static illumination and the mobile light separately. Both the mobile and static lights are captured only once; we model the light motion using rotation and falloff as described in Section~\ref{sec:illumination_model}.

\subsubsection{Model Initialization}
\label{sec:model_init}
Our method relies on an initial estimate of the eyeball pose and shape. As the subject is instructed to look at either the mobile camera or one of the static cameras, the initial eye pose can be estimated from the line-of-sight that connects the eyeball and camera centers.
We manually initialize eyeball pose and shape from three frames, roughly placing the eyeball in the correct position and shape in a 3D modeling software (Blender). Note that the initialization can be approximate, as it will be refined throughout the training process as shown in Fig.~\ref{fig:eyeball_geometry}.

\subsection{Network Training}
\label{sec:training}

\begin{figure}
    \centering
    \includegraphics[width=\columnwidth,page=2,trim={0 7.25cm 8cm 0},clip]{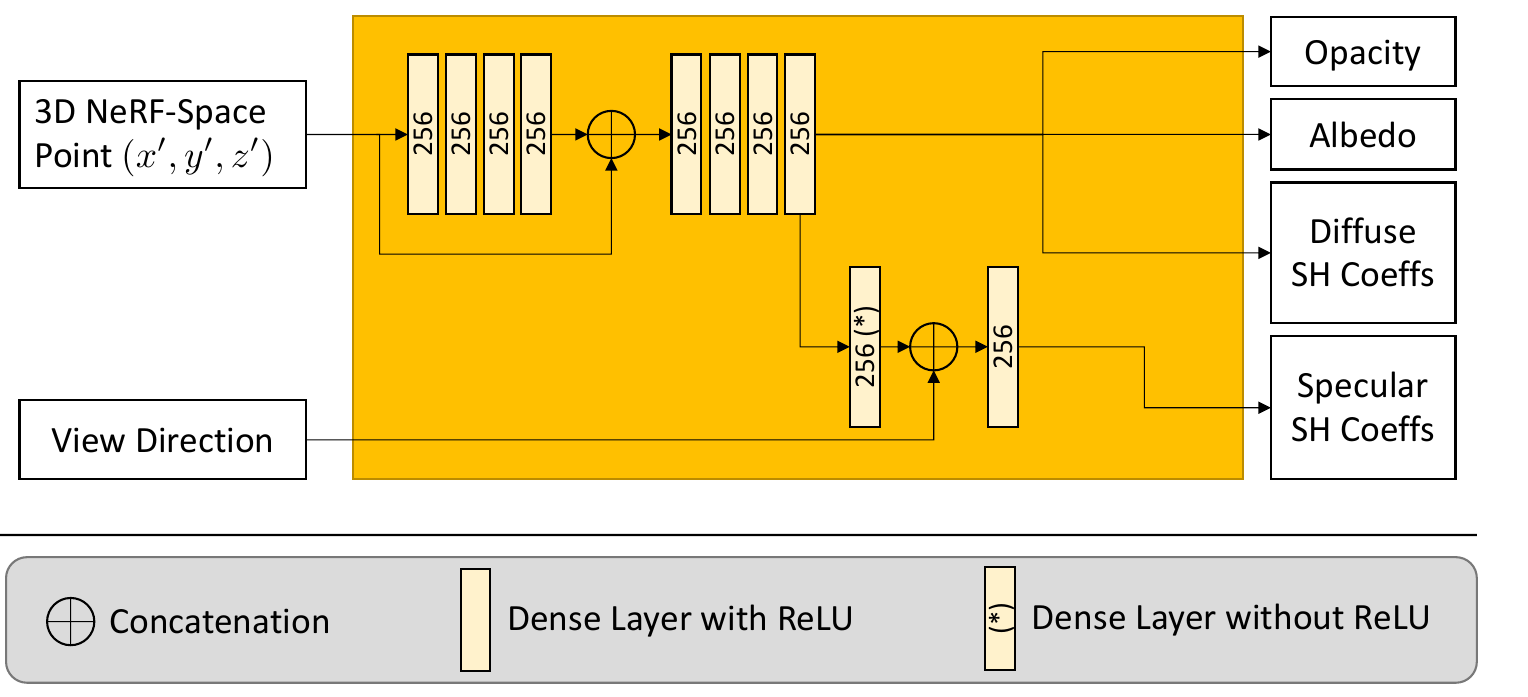}
    \caption{To improve initial network convergence, we start training with the simplified architecture shown above, and later on continue to train the full architecture as depicted in Fig.~\ref{fig:architecture}. The main difference is that here the diffuse SH coefficients only depend on the the 3D NeRF-Space points, where in the full model they also depend on the 3D World-Space points, in order to better model shadowing.}
    \label{fig:simple-architecture}
\end{figure}

As the main training loss, we use the mean squared error in sRGB space between the equidistant and importance sampled RGB values and the target pixel in the training image. 
The loss is computed on the outputs of the coarse and fine network
\begin{equation}
l_{\text{im}} =||\text{srgb}(x')-\text{srgb}(x_{f})||^2_2 + ||\text{srgb}(x'), \text{srgb}(x_{c})||^2_2 \: .
\end{equation}
As the sRGB transformation is not meaningful above 1, we use a linear transformation for such values. To encourage our hybrid model to represent specular reflection on the sclera by the explicit eyeball model, we ignore sclera pixels above a defined threshold when training the implicit volume.

In addition, we employ the non-negative SH loss for regularization. Each iteration, we randomly sample ten random directions, check if there are any negative predicted SH response functions, and then apply an $l_2$ loss on negative values. This avoids dead zones if the SH becomes negative. Furthermore, we observe that the diffuse shading is almost never non-positive anyway, and only apply this loss to the specular part of the shading. In order to reduce the uncertainty between diffuse and specular shading, we furthermore apply an $l_2$ loss to the specular coefficients. These losses are applied to the mean across all sample points.

\begin{equation}
\begin{split}
    l_{\text{noneg}} &= \mathbb{E}_{\wi\sim\Omega}[-\text{min}(0, \sum_{l=0}^{\text{order}} \sum_{m=-l}^{l}c_{lm}(\x, \wo)Y_{lm}(\wi))] \\
    l_{\text{spec}} &= \frac{1}{(\text{order}+1)^2}\sum_{l=0}^{\text{order}} \sum_{m=-l}^{l}||c_{lm}(\x, \wo)||_2^2
\end{split}
\end{equation}
We then apply an L2 loss on the per-vertex offsets ($N_V = 10242$), to avoid strong deviations from the underlying analytical model
\begin{equation}
    l_{\text{off}} = \sum_{i=0}^{N_V} \text{offset}_i^2 \:.
\end{equation}
Finally, we apply the same elastic regularization as used by Nerfies \cite{park2021nerfies} onto the warp field. We refer the reader to their paper on more details on how this loss term is computed.
This results in our final loss function
\begin{equation}
    l_{\text{tot}} = \lambda_{\text{im}} l_{\text{im}} + \lambda_{\text{noneg}} l_{\text{noneg}} + \lambda_{\text{spec}} l_{\text{spec}} + \lambda_{\text{elastic}}l_{\text{elastic}} + \lambda_{\text{off}} l_{\text{off}}
\end{equation}
where we use the empirically chosen weights $\lambda_{\text{im}} =$ 1, $\lambda_{\text{noneg}} =$ 1e-2, $\lambda_{\text{spec}} =$ 5e-4, $\lambda_{\text{elastic}} =$ 1e-3, $\lambda_{\text{off}} =$ 1e-6.

\begin{figure*}
    \centering
    \includegraphics[width=\textwidth,trim={0 0 8cm 0},clip]{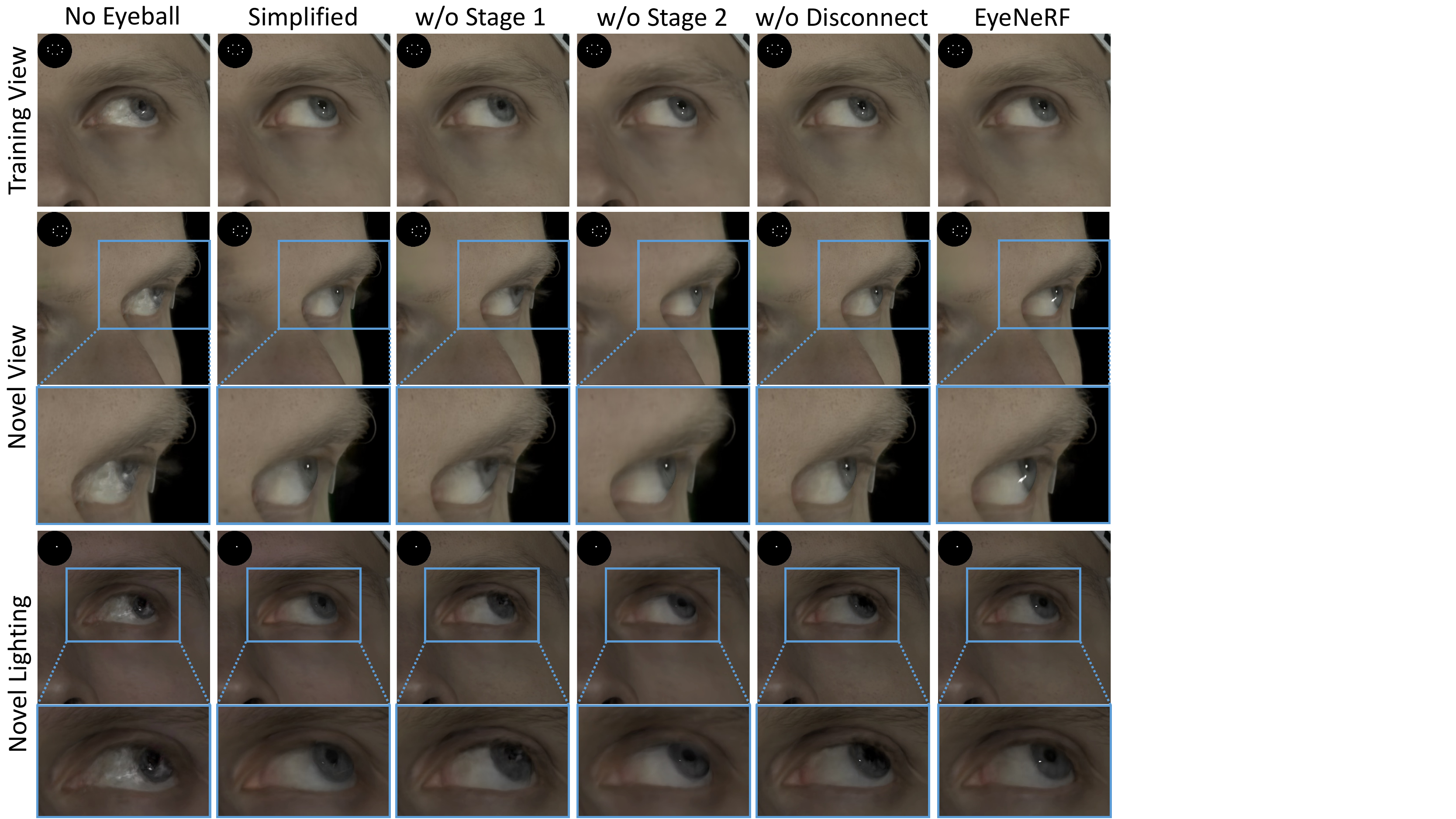}
    \caption{
    \textbf{Ablation Study.} We evaluate different design choices for \oursname. 
    Without the explicit eyeball model, the volumetric model fails to represent reflection and refraction, leading to strong artifacts.
    Without the first training stage, the eyeball pose and shape is inaccurate (see novel side view) hence many specular effects on the eyeball are not reproduced.
    Without the second training stage, the quality of the periocular region is reduced, yielding significant blur in the images.
    Results from the simplified architecture or without disconnect look similar to \oursname in the training view. However, they exhibit artifacts on the iris and sclera for novel views or novel illumination conditions. 
    }
    \label{fig:ablation}
\end{figure*}

We train \oursname in three stages, using slightly different architectures, on a total of 16 Nvidia V100s GPUs.
First, we use the simplified architecture for our \nerfname as shown in Fig.~\ref{fig:simple-architecture} to focus on learning the eyeball model parameters and per-frame rigid transformations. During this stage, every 50000 iterations, we additionally reset and reinitialize all learnable parameters other than the ones for our parametric model. We observe that this re-initialization greatly improves the final quality. This is due to two reasons. On the one hand, every time we reinitialize the volume, we remove all possible biases that may have been baked into the eyeball volume, for example due to view dependent effects. On the other hand, by resetting the volume we effectively blur it out, making the spatial gradients much smoother and therefore easier to learn with. For better signal when optimizing the eyeball pose and shape, we do not ignore the sclera pixels during this stage. This takes a total of approximately three days when trained on 8 of the 16 GPUs.
Second, we start training with our main \nerfname architecture as shown in Fig.~\ref{fig:architecture} to obtain initial network weights (for the warp field network and the \nerfname network) helping robustness of the third step. Note that this second step can take place in parallel with the first as they do not depend on each other, and also takes approximately three days, using the other 8 GPUs.
Third, once the eyeball model parameters and transformations have converged, we load them into our main architecture and continue training for roughly 100,000 iterations. During this final training stage, we disconnect the dotted connection in Fig.~\ref{fig:architecture} for points inside the eyeball by conditionally zeroing those values, to stronger condition the estimated SH lighting contribution on the world space points. We also disable the contribution from the specular shading, in order to encourage our network to model as much as possible using direct reflections. During this stage, we only use 8 GPUs, and train for roughly 24 hours.
As the first two stages are trained in parallel, our total pipeline can be trained in around four days.

\section{Results and Evaluation}
\label{sec:results}

\begin{figure}
    \centering
    \includegraphics[width=\columnwidth,trim={0 13cm 18.5cm 0},clip]{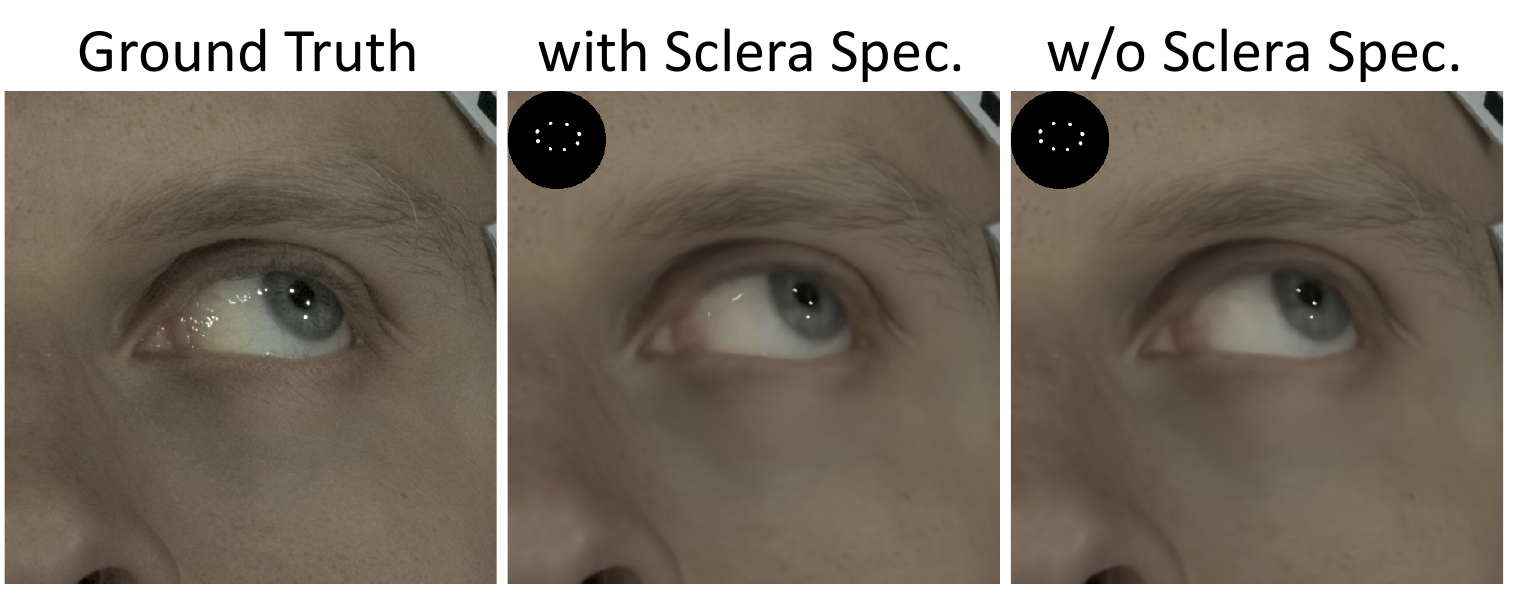}
    \caption{While the gross position of the highlights on the optimized sclera are correct, their appearance is visually off since the spatial resolution of our parametric eye model is not high enough to represent the high frequency surface variation caused by the lacrimal liquid and conjunctiva.}
    \label{fig:sclera_highlights}
\end{figure}

In this section, we discuss various results and evaluations of our method. We opted to suppress rendering of the specular highlights on the sclera for most results in this paper by using an IoR of zero for the sclera surface, because even though the gross position of the highlights on the optimized sclera are correct, their appearance is visually off since the spatial resolution of our parametric eye model is not high enough to represent the high frequency surface variation caused by the lacrimal liquid and conjunctiva (see Fig.~\ref{fig:sclera_highlights}).
We start with an ablation study (Sec.~\ref{sec:ablation}) to evaluate our design choices for \oursname.
Next, we compare to different baselines for gaze redirection and relighting (Sec.~\ref{sec:comparisons}).
Section~\ref{sec:additional_results} provides additional qualitative results such as eye animation, view synthesis, and relighting for different subjects as well as intrinsic decompositions into albedo and shading.  

\subsection{Ablation Study}
\label{sec:ablation}

\begin{table}
\caption{Results from our quantitative ablation study. Our full method outperforms or achieves comparable performance with the ablation settings on most metrics for all 3 tasks.}
\resizebox{\columnwidth}{!}{%
\begin{tabular}{ccccccc} 
      \toprule
      & \multicolumn{2}{c}{Novel View} & \multicolumn{2}{c}{Regazing} & \multicolumn{2}{c}{Relighting} \\
      \hline
       & MSE $\downarrow$ & SSIM $\uparrow$ & MSE $\downarrow$ & SSIM $\uparrow$ &  MSE $\downarrow$ & SSIM $\uparrow$ \\
      \hline
      No Eyeball & 1.10e-3 & 0.843 & 1.24e-3 & 0.838 & 9.62e-4 & 0.813\\ 

      Simplified &  9.74e-4  & 0.843 & 7.49e-4 & 0.853 & 7.53e-4 & 0.825\\ 

      w/o Stage 1 & 1.03e-3 & 0.847 & 1.07e-3 & 0.844 & 8.68e-4 & 0.822\\ 

      w/o Stage 2 &  1.78e-3 & 0.814 & 8.1e-4 & 0.835 & 7.58e-4 & 0.816\\ 

      w/o Disc.  &  7.78e-4 & 0.854 & \textbf{7.22e-4} & 0.855 & \textbf{7.19e-4} & 0.828\\ 

      EyeNeRF& \textbf{7.67-4} & \textbf{0.863} & 7.23e-4 & \textbf{0.857} & 7.23e-4 & \textbf{0.829}\\ 
      \bottomrule
\end{tabular}%
}
\label{tab:quant.eval}
\end{table}

We evaluate the following settings in our ablation study.
For network architectures and training stages, please refer to Section~\ref{sec:training}.
\begin{itemize}
    \item \textbf{No Eye Model:} We do not use the explicit eye model. There is no explicit refraction or reflection and all points are transformed using the warp field.
    \item \textbf{Simplified Architecture:} We only use the simplified architecture for \nerfname (Fig.~\ref{fig:simple-architecture}) and do not switch to the main branched architecture (Fig.~\ref{fig:architecture}). Everything is trained in a single training stage.
    \item \textbf{Without Stage 1:} We skip the first training stage in which the eyeball model parameters and rigid transforms are pretrained. All parameters are directly trained using the main branched architecture.
    \item \textbf{Without Stage 2:} We skip the second training stage in which the \nerfname and warp field weights are pretrained. Instead, the weights are trained from scratch using the already refined eyeball poses from the first training stage.
    \item \textbf{Without Disconnect:} Same as the final version but during the final training stage, the dotted connection in Fig.~\ref{fig:architecture} is not disconnected.
\end{itemize}
Fig.~\ref{fig:ablation} shows results of the ablation study. 
Please refer to our supplemental video for more visuals.
Without the eyeball model, the specular highlights on the eye need to be modelled by the volumetric representation. This leads to severe artifacts on the eyeball, especially for novel views or novel lighting scenarios.
Without the first training stage, the eyeball pose and shape is inaccurate as shown in the novel side view. Due to the incorrectly placed eyeball, the specular reflections are missing in many cases.
Without the second training stage, the quality of the periocular region degrades significantly which is especially visible in areas with high-frequency details such as the eyebrows.
Whereas the results with the simplified architecture or without disconnect look similar to \oursname from the training views, the iris and sclera quality reduces significantly for novel views or novel lighting.

We show results from the quantitative ablation study for novel view synthesis, regazing and relighting in Table~\ref{tab:quant.eval}. 
For novel view synthesis, we select a number of frames, from which we exclude one of the four camera views during training as a held-out test view.

For both regazing and relighting, we define a set of held-out test frames, for which all camera views are excluded from training. For regazing, we use frames from the sequence where the captured subject holds their head still and looks at the moving camera. For relighting, we use the sequence where only a moving light is used for illumination. However, we note that our initial estimates for gaze (see Section~\ref{sec:model_init}) are not completely accurate. 
For the test frames, we optimize the eyeball pose using the same strategy as for training frames. We use the so-computed eyeball pose for synthesizing the test frames using our trained model and perform quantitative comparison with the ground-truth to obtain the results reported in Table~\ref{tab:quant.eval}.

We compute SSIM and MSE as quantitative image comparison metrics for models trained for all settings in the ablation study. As our ablation study conditions primarily differ in how the eye is modeled, we use a smaller 300x300 crop around the eye while computing the metric, in order to more strongly focus on the eye itself.
Please note that these image metrics have their limitations, e.g. over-penalizing misalignments or under-penalizing blur. Nonetheless, the results indicate that our full method outperforms or is on par with the other ablation study conditions.

\subsection{Comparisons}
\label{sec:comparisons}

We compare \oursname to existing methods for regazing \cite{he2019gazeredirection} and relighting \cite{Pandey:2021}.

\begin{figure}
    \centering
    \includegraphics[draft=false,width=\columnwidth,trim={0 12.11cm 9.36cm 0},clip]{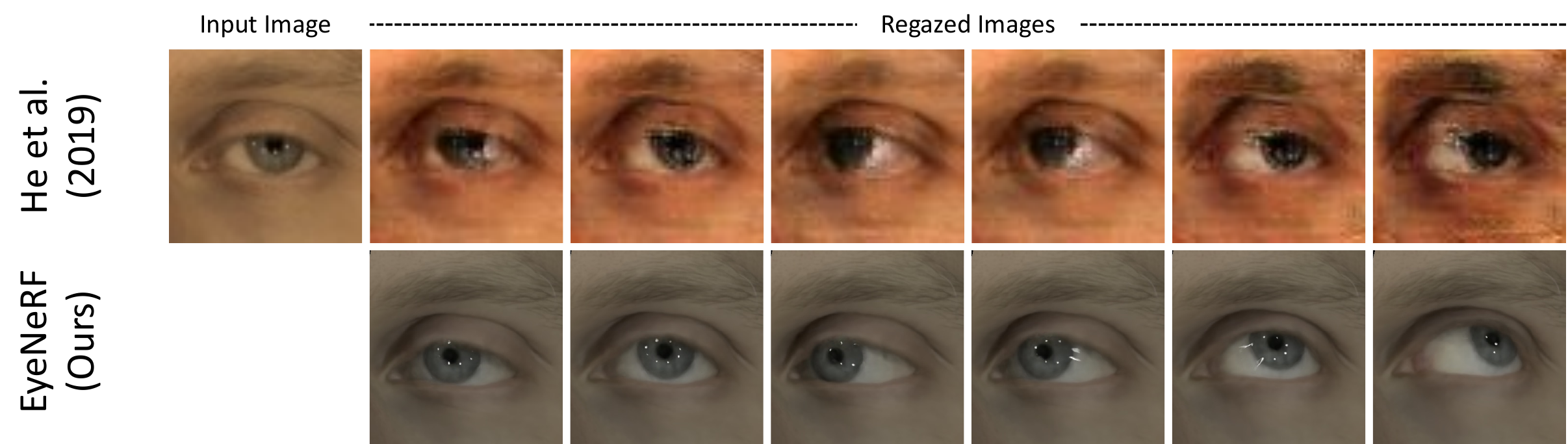}
    \caption{We compare with \cite{he2019gazeredirection} for regazing. \oursname{} controls the gaze more accurately and preserves the identity better. Please note that \cite{he2019gazeredirection} operates on a single input image only, where ours is a much more involved setting.}
    \label{fig:comparison_regazing}
\end{figure}

\begin{figure}
    \centering
    \includegraphics[draft=false,width=\columnwidth,clip]{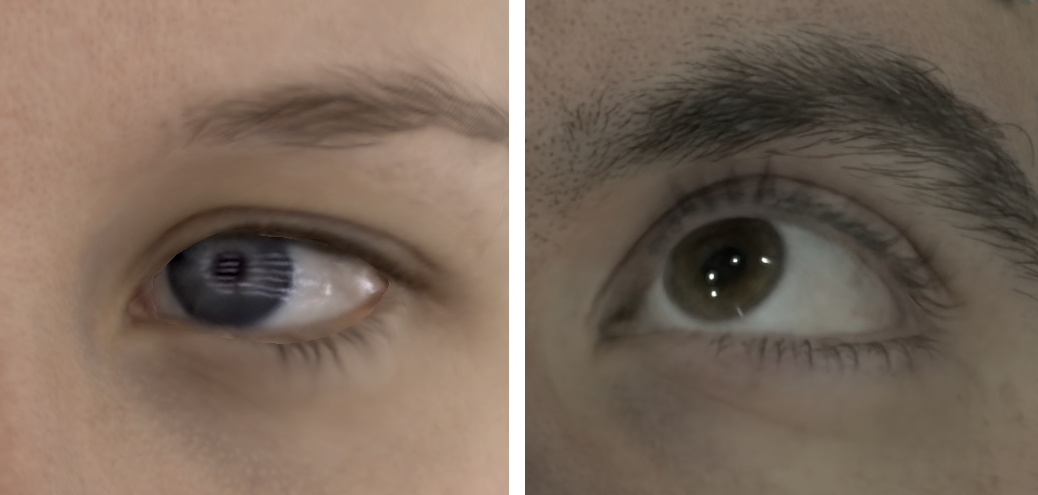}
    \caption{
    We compare an image synthesized by our method (right) with an image generated by \citet{schwartz2020eyes} [\copyright Schwartz et al.] (left). While \citet{schwartz2020eyes} show that they can perform view synthesis and regazing, their method does not allow for capturing high frequency skin/eye reflectance, perform relighting or capturing and synthesizing thin structures such as eyelashes and eyebrows in 3D.
    }
    \label{fig:comparison_regazing_schwartz}
\end{figure}

\begin{figure}
    \centering
    \includegraphics[draft=false,width=\columnwidth,clip]{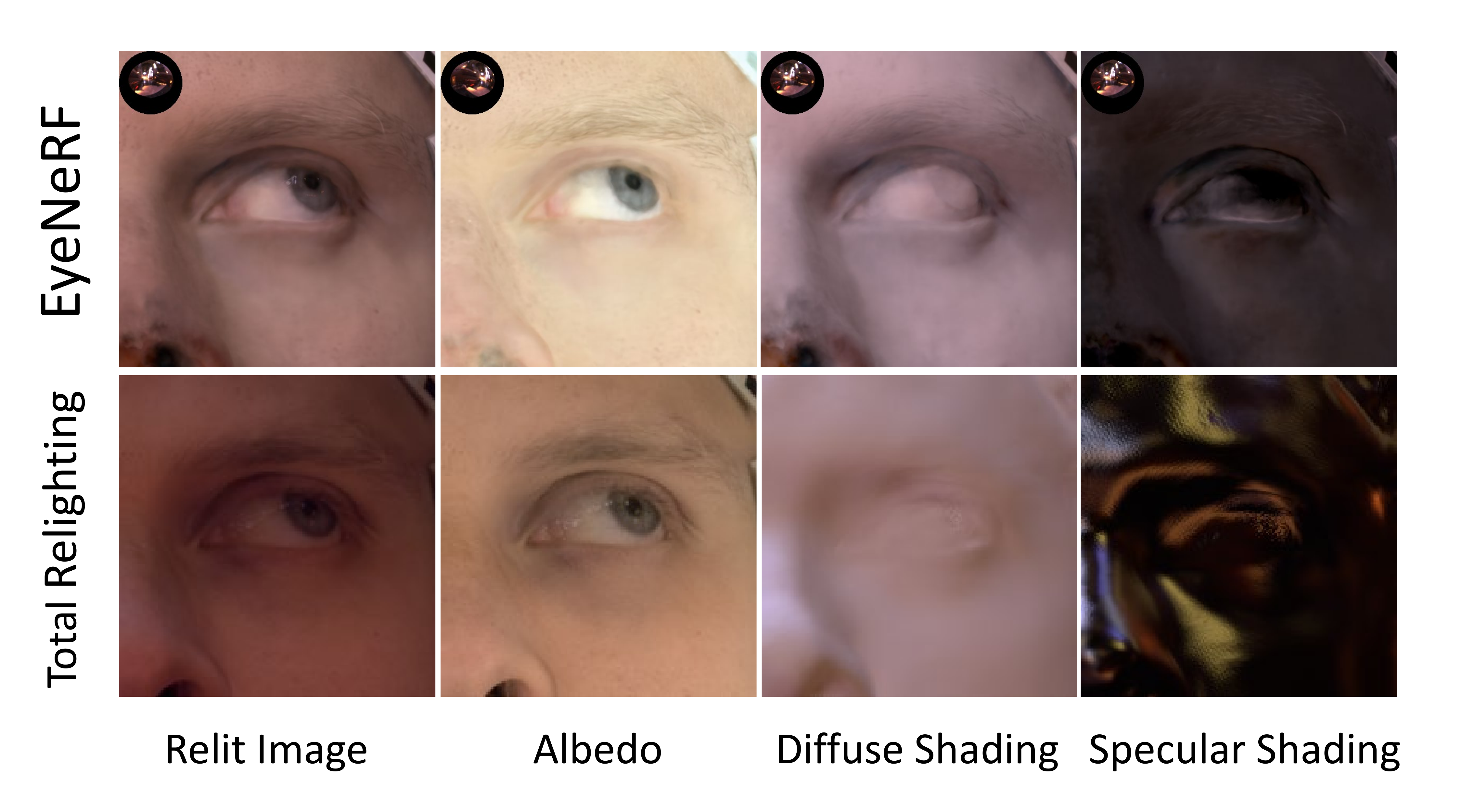}
    \caption{We compare with \citet{Pandey:2021} for relighting. \oursname{} produces higher quality relighting, particularly the environmental reflections on the eye.}
    \label{fig:comparison_relighting}
\end{figure}

\paragraph{Regazing}
We perform regazing on subject 1 and compare with \citet{he2019gazeredirection} in Fig.~\ref{fig:comparison_regazing}. \oursname can synthesize more extreme gaze angles, it preserves the identity better, and shows less artifacts. Note that their method is defined for a different setting, as it can be applied for a single image, and works across multiple identities. Our setting is much closer to that of \citet{schwartz2020eyes}, who also use multi-view footage of the same subject for regazing. Unfortunately a direct comparison with \citet{schwartz2020eyes} is difficult because their data capture setting uses a dense multi-view setup, and their implementation is not available publicly or on request. Nonetheless, we show a demonstrative comparison in  Fig.~\ref{fig:comparison_regazing_schwartz}. Our method enables additional capabilities such as high-quality relighting and 3D capture of thin structures associated with the eye region.

\paragraph{Relighting}
We compare environmental relighting on a single frame against the state-of-the-art portrait relighting method of \citet{Pandey:2021} in Fig.~\ref{fig:comparison_relighting}. They use high-quality dense Light Stage data to train their technique. While they do estimate surface normals, they do not estimate the full 3D geometry of the face, hence they are unable to model the full light transport. \oursname{} can synthesize very high quality relighting, including self-shadowing, particularly in the eye region.

\subsection{Additional Results}
\label{sec:additional_results}
\begin{figure}
    \centering
    \includegraphics[width=\columnwidth]{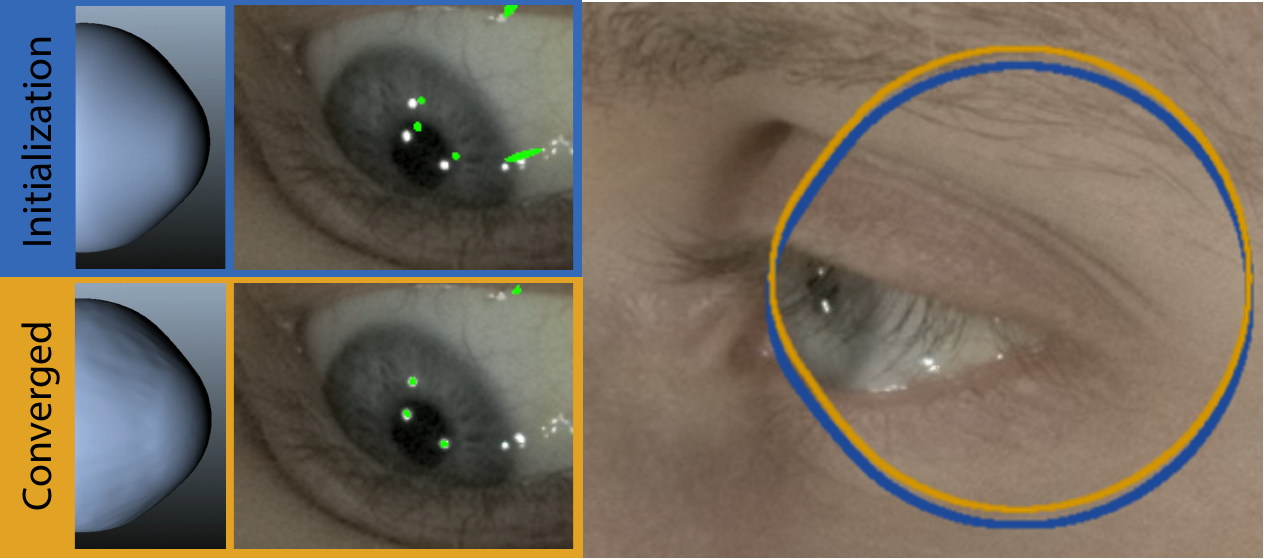}
    \caption{The initial estimate (blue) for the eyeball pose and shape is refined (orange) during training (Sec.~\ref{sec:training}). \textbf{left row:} the shape changes and the sclera becomes bumpy while the cornea remains smooth. middle row: the synthesized highlights (green) are initially off but coincide with the ground-truth highlights for the converged shape. \textbf{right:} eyeball contours of subject 1 at the beginning (blue) and at the end of training (orange).}
    \label{fig:eyeball_geometry}
\end{figure}

\begin{figure*}
    \centering
    \includegraphics[page=3, width=0.92\textwidth, trim={0cm 4.45cm 0cm 0cm}, clip]{figures/fig_allsubjects.pdf}
    \includegraphics[page=1, width=0.92\textwidth, trim={0cm 4.45cm 0cm 0.72cm}, clip]{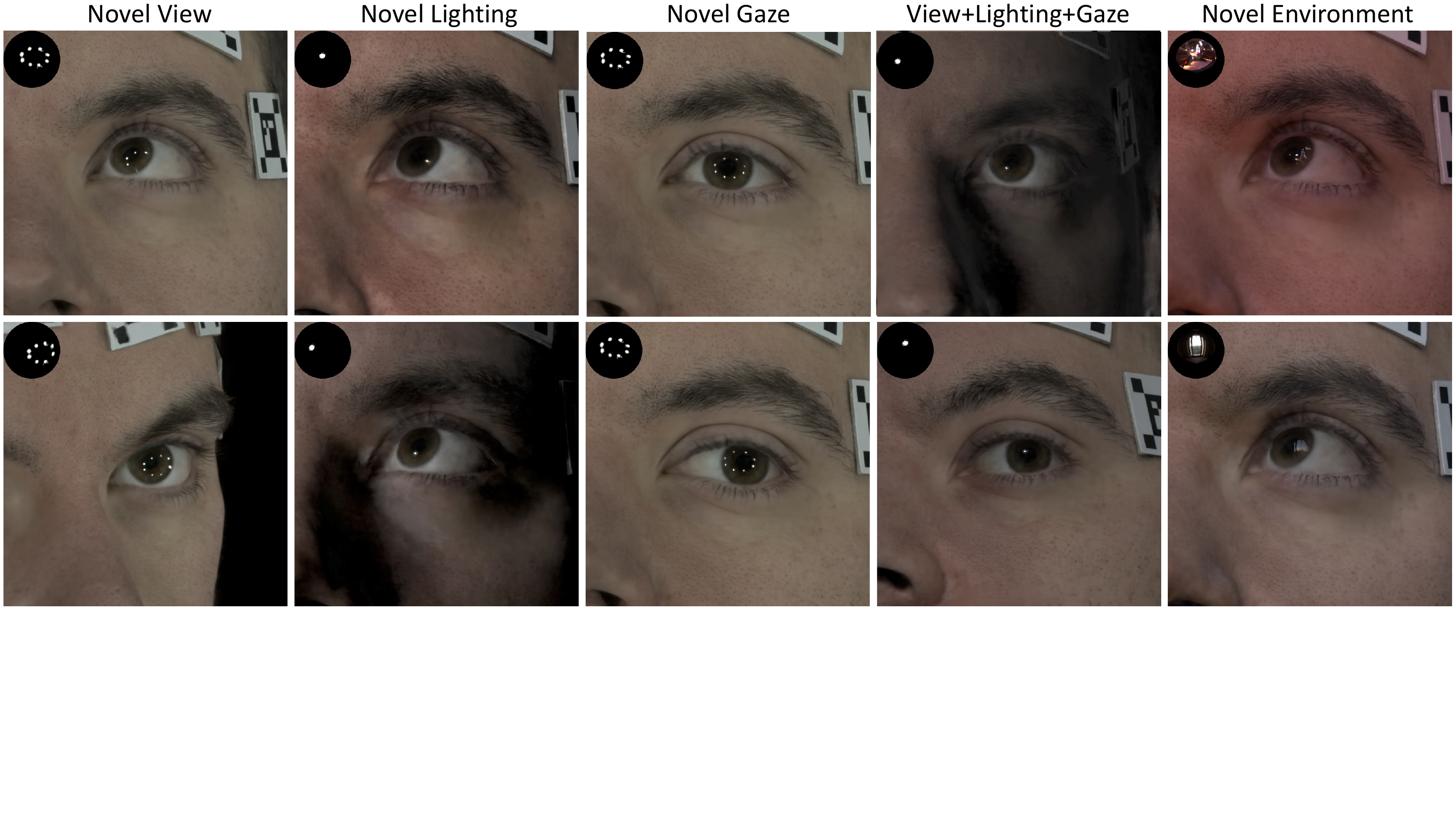}
    \includegraphics[page=2, width=0.92\textwidth, trim={0cm 4.45cm 0cm 0.72cm}, clip]{figures/fig_allsubjects.pdf}
    \caption{Here we show synthesis results for different subjects. In the three left columns, we vary only one attribute (view, lighting, or gaze) from a training image. However, \oursname can also generate combinations of unseen attributes (4th column) or use arbitrary environment maps for relighting (5th column).}
    \label{fig:qualitative}
\end{figure*}

\begin{figure*}
    \centering
    \includegraphics[width=\textwidth,trim={0 12.61cm 0 0},clip]{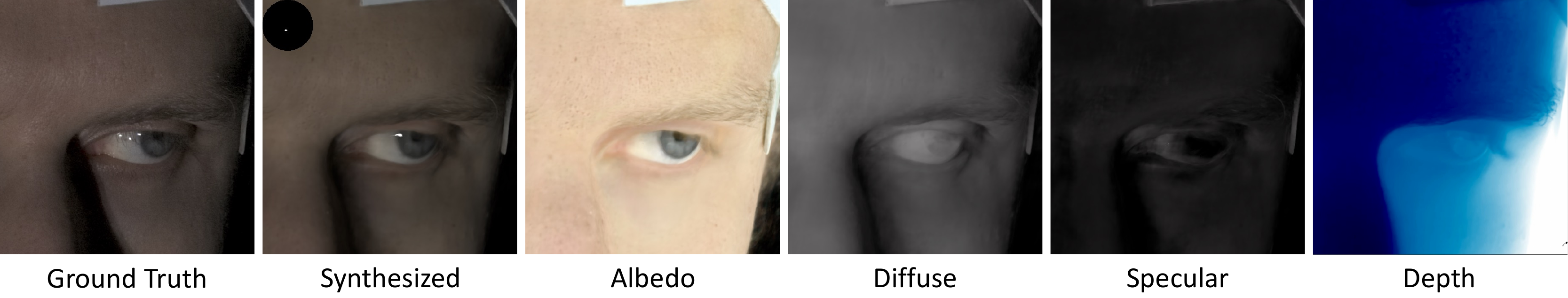}
    \caption{We show a side-by-side comparison with a ground-truth image that has not been seen during training (first two columns). The third to sixth columns show the corresponding intrinsic decomposition.}
    \label{fig:validation_vs_gt}
\end{figure*}

\begin{figure*}
    \centering
    \includegraphics[width=\textwidth, trim={0 0 4.25cm 0}, clip]{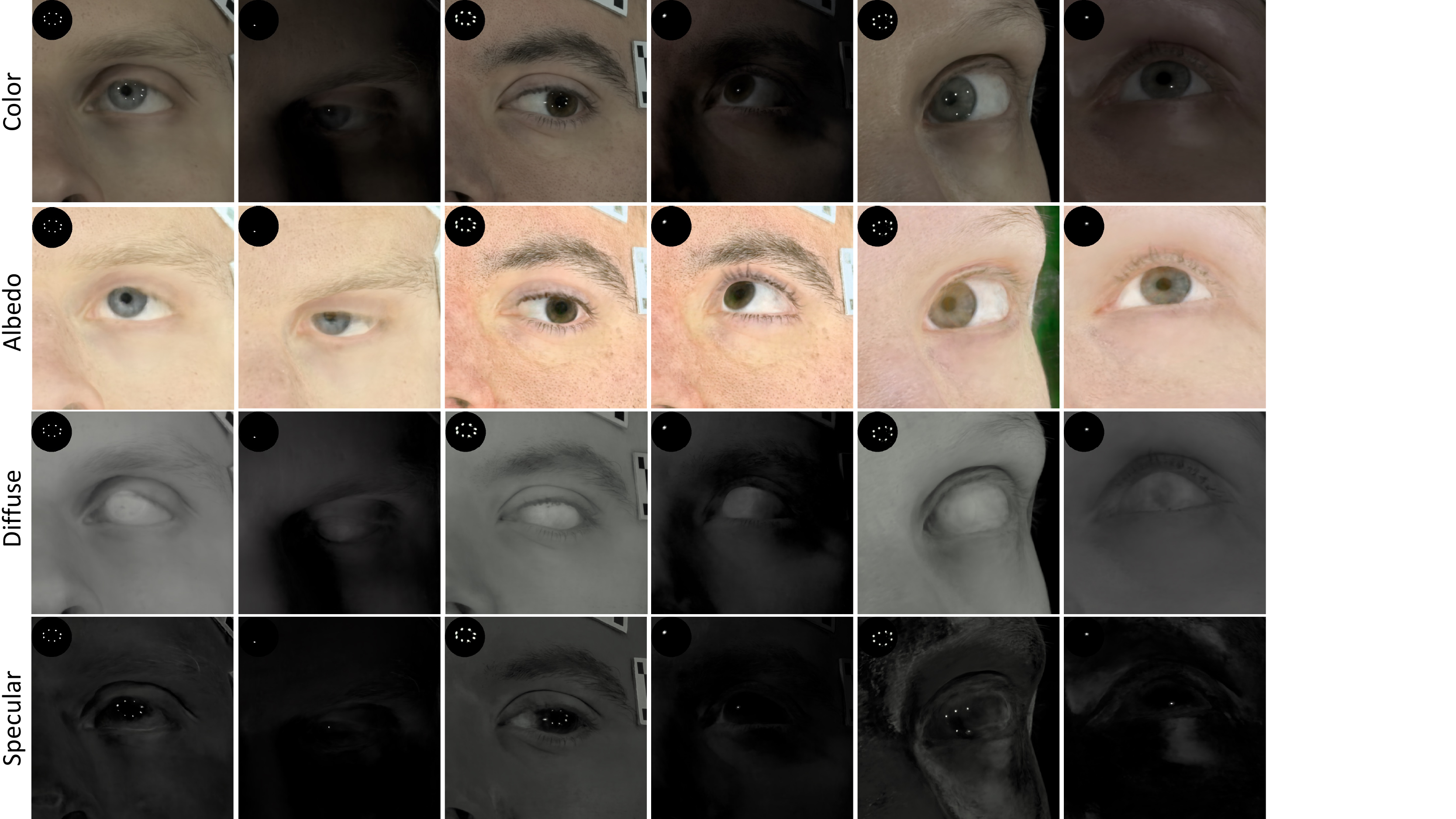}
    \caption{Intrinsic decomposition results for different subjects. \oursname is able to disentangle albedo, diffuse and specular shading.}
    \label{fig:decomposition}
\end{figure*}

\paragraph{Eyeball Model Refinement.}
In Fig.~\ref{fig:eyeball_geometry}, we demonstrate how \oursname successfully refines the eyeball geometry during the course of training.
This leads to more accurately synthesized highlights and a better aligned eyeball contour.
\paragraph{Synthesizing View, Lighting, and Gaze.}
In Fig.~\ref{fig:validation_vs_gt}, we compare our synthesized result to an unseen ground-truth image.
We show additional results in Fig.~\ref{fig:qualitative}. 
For each subject, we show results for deviating from a training frame by changing a single property (view, lighting condition, or gaze directions). 
Furthermore, \oursname also synthesizing compelling results for combinations of unseen scene properties (view, lighting, and gaze) or relighting using an environment map.
\paragraph{Intrinsic Decomposition.}
In Fig.~\ref{fig:decomposition}, we show the final rendered image for different subjects as well as decompositions into albedo, diffuse and specular shading. 

\section{Discussion \& Limitations}
\label{sec:discussion}

Although we demonstrate exceptional overall quality, our method does still have a number of shortcomings and limitations.

One major drawback is the requirement of multiview, multi-lighting, multi-gaze data. As previously mentioned, very few (if any) datasets exist which contain all of the above. Furthermore, capturing this kind of data, although lightweight compared to many professional capture setups such as a full light stage, still requires very high quality cameras and good lighting. Additionally, as we capture the subjects under a variety of conditions, we require around 40 minutes per subject. To reduce complexity we rely on AR tags to track the rigid pose, which requires attaching small tags to the subjects face. Relying instead on a face tracker would be a more convenient alternative.

Another drawback is the long training and evaluation time. Our full pipeline takes around 4 days to train on 8 Nvidia V100s, and currently requires about thirty seconds to render an 800x800 sized image. The majority of the computational cost lies in the NeRF-SHL and NeRFies networks and recently there have been a surge of papers that greatly accelerate training such networks (e.g. \cite{mueller2022instant}).

A more fundamental limitation of our method is the inability to model high frequency effects other than the cornea reflection. NeRF and NeRF-like methods often already struggle with exceptionally high frequency details, such as with the eyelashes, especially when time-varying due to deformation. A lot of the detail loss can be attributed to the deformation field, which struggles to accurately warp the frames to the canonical volume, e.g. for the upper eyelid.

By approximating the light transport equation using spherical harmonics, we are inherently limited by the frequency of the SH degree we choose to use. As such, we are unable to model very high frequency effects, such as strongly specular reflections on the skin. As we model the entire light transport function using these functions, this further implies that we have similar issues with high frequency shadows for example. Since we designed our method for human faces these problems are less of an issue, but we do not expect \oursname to work nearly as well in a more general setting. Going to higher order SH will quickly become intractable as the number of coefficients grows in a squared manner. A potential solution could be to pool information spatially across a patch of skin and thus provide more constraints to estimate the coefficients, or to explore alternative reflectance representations (i.e. data driven priors).

Furthermore, we are currently unable to reconstruct the high-frequency sclera reflections since the surface resolution of our eyeball model is not high enough. Increasing the resolution however will only be beneficial up to a certain amount since learning high-frequency spatial variation in the current setting is also limited by the image resolution. At the current image resolution pixels are either specular or not, which does not provide good gradients for learning. In addition we are currently not modeling the relative deformation of the conjunctiva and the sclera, which leads to blurred veins, as well as the pupil dilation. Exploring deformation fields within the eye itself could potentially address this.

Finally, we currently do not model expressions (lid closing, squinting, smiling, etc), and assume the periocular shape is entirely correlated to eye gaze. Additionally capturing expression and properly disentangling gaze and expression would be an exciting avenue for future work.


\section{Conclusion}
\label{sec:conclusion}

We presented \oursname, a novel hybrid model combining the strengths of explicit parametric surface models with implicit neural representations, specifically designed to represent the greatly varying visual properties of the periocular region. Our model enables explicit control over the gaze and to render realistic images from novel viewpoints under novel illumination. \oursname is well suited to generate visually rich training data of the periocular area to train ML models, i.e. for gaze prediction, or to create visually compelling content for VFX productions. While our method has several limitations (see Sec.~\ref{sec:discussion}) which inspire future work, it is a leap forward towards realistic reproduction of one of the most challenging and most important areas of the human face.

\FloatBarrier

\begin{acks}
\begingroup

\setlength{\columnsep}{10pt}%
\setlength\intextsep{0pt}
This project has received funding from the European Research Coun\-%
\begin{wrapfigure}[3]{r}{0.38\columnwidth}
\centering
 \includegraphics[width=0.38\columnwidth]{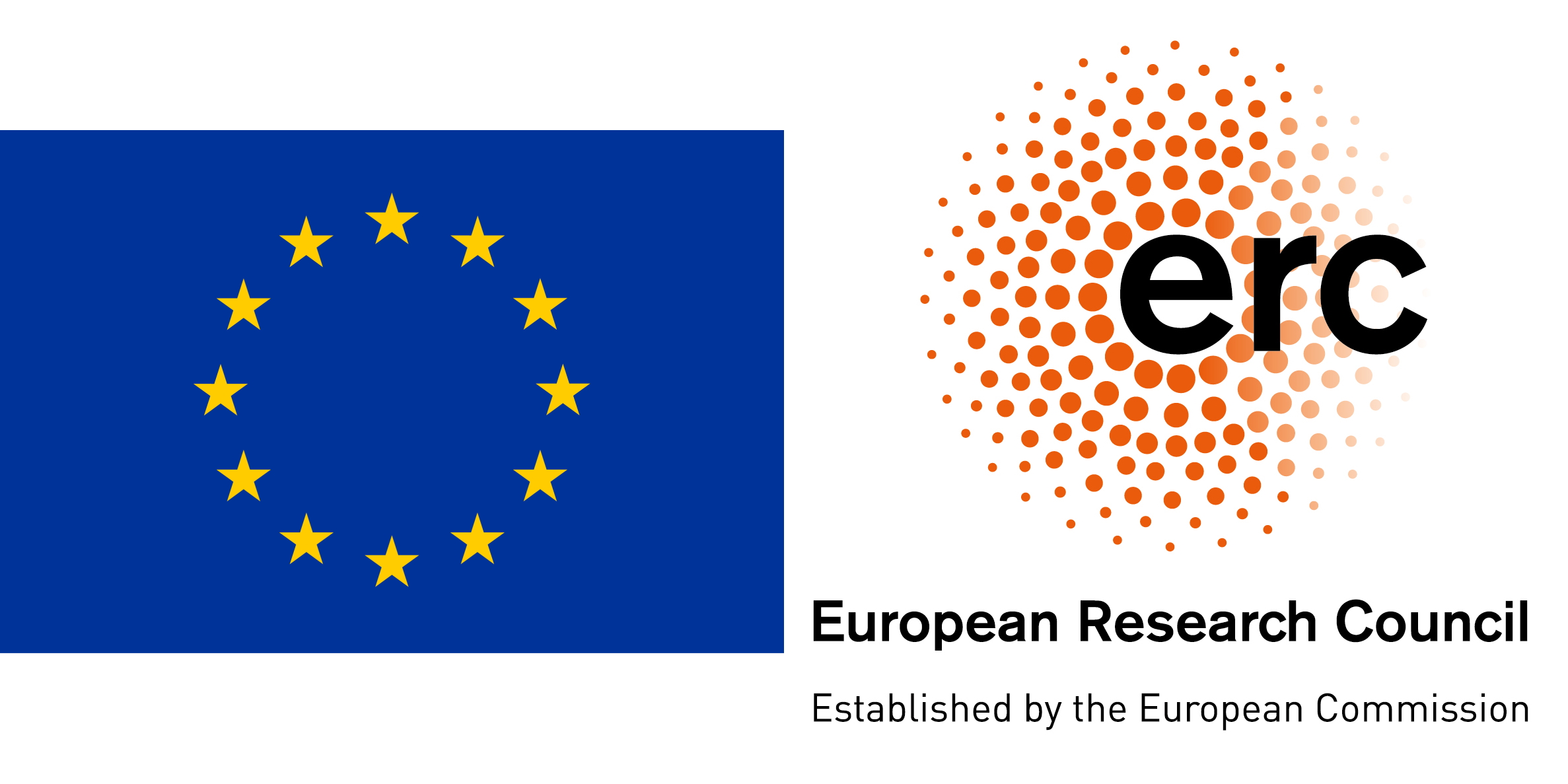}
\end{wrapfigure}%
cil (ERC) under the European Union’s Horizon 2020 research and innovation programme grant agreement No 717054.

\endgroup
\end{acks}

\bibliographystyle{ACM-Reference-Format}
\bibliography{references}

\appendix
\section{SH Environment Map Rotation}
\label{app:SH_rotation}

Rotating the environment map can be done in the spherical harmonics domain extremely efficiently.
In particular, for any given set of SH coefficients $c$ (as a vector), we can compute $R'$ as a block diagonal with blocks of size (1x1, 3x3, 5x5, etc.\:equal to the corresponding SH order), from any given rotation matrix $R$. We compute $R'$ using the method shown by \citet{ivanic1996rotation}.

For the purpose of simplicity, we assume that the spherical harmonics coefficients and functions of order $l$ and degree $m$ are stacked into a vector.
Given the matrix $R'$ computed from $R$, we have the following property
\begin{equation}
    \mathbf{Y}(R(\omega)) = R'(\mathbf{Y}(\omega)) \: .
\end{equation}
We note that to rotate an environment map, we need to query it at the inverse rotated locations
\begin{equation}
    R(E)(\omega) = E(R^{-1}(\omega) \: ,
\end{equation}
resulting in the following equation
\begin{equation}
    L_o(x, \omega_o) \approx \int_{\Omega} \sum_{l=0}^{\text{order}} \sum_{m=-l}^{l}c_{lm}(x, \wo)Y_{lm}(\wi) E(R^{-1}(\wi)) d \wi \:.
\end{equation}
Then, we rotate both $\wi$'s simultaneously by $R$ by change of variables. 
In particular, rotating the integral domain does nothing, and the determinant of the Jacobian is 1 as it is a rotation.
\begin{equation}
   L_o(x, \wo) \approx \int_{\Omega} \sum_{l=0}^{\text{order}} \sum_{m=-l}^{l}c_{lm}(x, \wo)Y_{lm}(R(\wi)) E(\wi) d \wi 
\end{equation}
We do the same reordering as in Sec.~\ref{sec:env_map}, and reinterpret the spherical harmonics functions as the vector $\mathbf{Y}$ 
\begin{equation}
    L_o(x, \wo) \approx \sum_{l=0}^{\text{order}} \sum_{m=-l}^{l}c_{lm}(x, \wo) \left(\int_{S\Omega}\textbf{Y}(R(\wi)) E(\wi) d \wi\right)_{lm} \: .
\end{equation}
We can then use our previously stated SH rotation property
\begin{equation}
L_o(x, \wo) \approx \sum_{l=0}^{\text{order}} \sum_{m=-l}^{l}c_{lm}(x, \wo) \left(\int_{\Omega}R'(\textbf{Y}(\wi)) E(\wi) d \wi\right)_{lm} \: .
\end{equation}
Using the fact that $E(\wi)$ is a scalar, we can then reorder the equation into the following form
\begin{equation}
    L_o(x, \wo) \approx \sum_{l=0}^{\text{order}} \sum_{m=-l}^{l}c_{lm}(x, \wo) R'\left(\int_{\Omega}\textbf{Y}(\wi) E(\wi) d \wi\right)_{lm} \:.
\end{equation}
We observe that this is only a slight deviation from the original formula. Now we merely need to apply a single matrix multiplication to the previously precomputed integral.

\section{Eye Model Parametrization}
\label{app:parametrization}

As mentioned in Section~\ref{sec:expliciteyemodel}, we use a model very similar to \citet{schwartz2020eyes}, with only one difference. For their eye model, they warp a sphere into their desired shape, by using a smoothstep function conditioned on the angle of the original sphere relative to the eye gaze direction. In particular, they define the linear blending parameter as:
\begin{equation}
    \alpha = \text{smoothstep}(2\theta_{\text{diff}} + 0.5)
\end{equation}
We replace 2 and 0.5 using learnable coefficients $\theta_{\text{mod}}$ and $\theta_{\text{offset}}$, which are constrained to lie between 1 and 3, and -0.5 and 1.5 respectively, using a tanh function.
For further details, we refer the reader to the appendix of \citet{schwartz2020eyes} where they describe their model in full detail.

\section{Optimization Details}
\label{app:optimization details}
In order to train the warp field more efficiently, we use annealing on the positional encoding as proposed by Nerfies.
We add a constant offset to the sample points inside the eyeball after the rigid transformation to ensure that the rigidly transformed eyeball interior does not overlap with the deformed periocular region volume. 
Furthermore, in order to allow the network to correct for cases where the eyeball volume is misaligned with the eyeball model surface, we learn a global 6-DoF transformation on top of the transformation per frame, which does not affect the volume. This greatly improves training performance when there is some initial bias.

As with Nerfies and NeRF, we use a separate coarse and fine MLP with approximately 700k optimizable parameters each (for the full NeRF-SHL network). The deformation MLP has appxorimately 100k optimizable parameters, plus an 8-dimensional encoding for each frame. As explained in Section \ref{app:parametrization}, our eye model has 5 shape parameters. We then have 6 parameters which encode a global rigid transformation, plus 6 parameters per frame for the per-frame transformation. Finally, we learn a scalar environment map scale factor.
We use the Adam optimizer to train our models, with a learning rate of 0.001 which decays exponentially by a factor of 10 every 100'000 timesteps. We use a batch size of 2816 rays, with 192 coarse and 192 fine samples each. 
In order to improve training performance, we learn the parameters at a different scale, by premultiplying the underlying values with a constant factor. 
As Adam is invariant to diagonal rescaling of gradients, this effectively corresponds to different learning rates for each parameter. 
We use the following empirically derived premultiplication factors: 

\begin{tabular}{|c|c|c|c|c|c|}
    \hline
    glob. trans & glob. rot & trans & rot & v. offsets & eye model\\
    \hline
    0.0005 & 0.05 & 0.0001 & 0.01 & 0.00005 & 0.0005\\
    \hline
\end{tabular}

\section{Capture Details and Image Preprocessing}

For each capture condition, we record between 1.5 to 3 minutes of footage at 4k resolution with 24 fps. In order to have tolerance for head motion, we use an f-stop value of 9, an ISO of 1000 and an exposure time of 1/48 seconds, and focus the camera to the subjects' right eyes in the neutral pose. We use ProRes 422 compression for improved quality.
In order to temporally synchronize the static cameras and mobile camera, we turn a cell phone light on and off briefly, and sync on the frame where it turns off. We then extract only every 8-th frame to reduce redundancy and the amount of data that needs to be handled.
We only aim to reproduce the eye and periocular region. As such, we crop the original image to 800x800 centered around the eye. To do so, we use the projected eyeball center estimate as the center of the image crop. This also allows to estimate a near and far culling plane, which we define as 10 cm in front of and behind the eyeball center.
We then discard the frames with bad or nonexistent AR tag detection, for example due to motion blur or poor illumination quality. 
Finally, to minimize the data handling, we randomly sample a set of frames from each capture condition for training. We use 100 frames from the sequence where the subject follows the mobile camera with their gaze. For the three other conditions, we sample 40 frames each from the sequences where they look at each of the four static cameras, resulting in a total of 160 frames each. This results in a total of 580 training frames for each of the 4 camera views.

\end{document}